\documentclass[9pt,shortpaper,twoside,web]{ieeecolor}
\usepackage{generic}
\usepackage{cite}
\usepackage{amsmath,amssymb,amsfonts}
\usepackage{graphicx}
\usepackage{textcomp}
\usepackage{subfigure}
\usepackage{float}
\usepackage{graphicx}
\usepackage{comment}
\usepackage{extpfeil}
\usepackage{wrapfig}

\usepackage{color}
\usepackage{amssymb}
\graphicspath{{./figures/}}

\begin{document}

\title{Lesion Segmentation in Ultrasound Using Semi-pixel-wise Cycle Generative Adversarial Nets}

\author{ Jie~Xing, Zheren~Li, Biyuan Wang, Yuji Qi, Bingbin Yu, Farhad~G.~Zanjani,\\
       Aiwen~Zheng, Remco~Duits, Tao~Tan 
\thanks{Jie~Xing, Zheren~Li have equal contribution }% <-this % stops a space
\thanks{J. Xing and Ai-Wen Zheng are with Zhejiang cancer hospital, China }
\thanks{Z. Li is with School of Biomedical Engineering, Shanghai Jiao Tong University, Shanghai, China}
\thanks{T. Tan and R. Duits are with Department of Mathematics and Computer Science,
Eindhoven University of Technology, Eindhoven 5600 MB, The Netherlands }% <-this % stops a space
\thanks{Farhad~G.~Zanjani is with Department of Electrical Engineering,
Eindhoven University of Technology, Eindhoven 5600 MB, The Netherlands }% <-this % stops a space
% <-this % stops a space
\thanks{B. Wang is with Department of Computing, Tokyo Institute of Technology, Tokyo, Japan}
\thanks{Y. Qi is with Department of Biomedical Engineering, Yale University, New Haven, USA}
\thanks{B.Yu is with Robotic Innovation Center, German Research Center of Artificial Intelligence, Bremen, Germany}
\thanks{T. Tan (t.tan1@tue.nl) is the corresponding author}
}

%\markboth{Journal of \LaTeX\ Class Files,~Vol.~13, No.~9, September~2014}%
%{Shell \MakeLowercase{\textit{Tan et al.}}: Lesion segmentation in ultrasound}

\maketitle

\begin{abstract}
Breast cancer is the most common invasive cancer with the highest cancer occurrence in females. Handheld ultrasound is one of the most efficient ways to identify and diagnose the breast cancer. The area and the shape information of a lesion is very helpful for clinicians to make diagnostic decisions. In this study we propose a new deep-learning scheme, semi-pixel-wise cycle generative adversarial net (SPCGAN) for segmenting the lesion in 2D ultrasound. The method takes the advantage of a fully convolutional neural network (FCN) and a generative adversarial net to segment a lesion by using prior knowledge. We compared the proposed method to a fully connected neural network and 
% * <farhad.ghazvinian@gmail.com> 2018-12-14T07:29:40.051Z:
% 
% > fully connected neural network
% fully convolutional networks
% 
% ^.
% * <farhad.ghazvinian@gmail.com> 2018-12-14T07:28:36.626Z:
% 
% > fully connected convolutional
% Perhaps you mean "fully convolutional networks"
% 
% ^ <farhad.ghazvinian@gmail.com> 2018-12-14T07:29:14.308Z.
the level set segmentation method on a test dataset consisting of 32 malignant lesions and 109 benign lesions. 
Our proposed method achieved a Dice similarity coefficient (DSC) of 0.92 while FCN and the level set achieved 0.90 and 0.79 respectively. Particularly, for malignant lesions, our method increases the DSC (0.90) of the fully connected neural network to 0.93 significantly (p$<$0.001). The results show that our SPCGAN can obtain robust segmentation results. The framework of SPCGAN is particularly effective when sufficient training samples are not available compared to FCN. Our proposed method may be used to relieve the radiologists' burden for annotation.
% * <farhad.ghazvinian@gmail.com> 2018-12-14T07:30:37.235Z:
% 
% > fully connected neural network
% !
% 
% ^ <farhad.ghazvinian@gmail.com> 2018-12-14T07:31:02.959Z.

\end{abstract}

\begin{IEEEkeywords}
Lesion Segmentation, Deep Learning, Generative Adversarial Networks, Breast Cancer, Ultrasound Image Analysis
% * <farhad.ghazvinian@gmail.com> 2018-12-14T07:31:42.372Z:
% 
% > Deep Learning, Generative Adversarial Networks, Breast Cancer, Ultrasound Image Analysis
% perhaps should be written in small letters
\end{IEEEkeywords}

% \linenumbers

\section{Introduction}
\label{}

Breast cancer is one of the leading causes of death for women in the UK. According to the statistics published by Cancer Research UK, there are about 155 women in 100,000 suffering from breast cancer in the UK and incidence rate is around 10\% for females in other European countries while this number is over 12.5\% for breast cancer with the American females\cite{DeSantis2015}. Since the causes of breast cancer still remain unknown, early diagnosis of breast cancer plays a significant role in reducing the death rate and maintaining the quality of the life after treatments\cite{Kaul2002}.

Ultrasound imaging technology has developed rapidly in recent years. Compared to mammography, there is no radiation damage to women from ultrasound imaging. It is easy to obtain any cross-sectional images of breast tissue by manipulating the handheld ultrasound while normally only two projections are obtained from mammography. It provides an easy way to assess if a lesion is solid or fluid-filled\cite{Laine1995}. Ultrasound detects early-stage cancers in women with mammography-negative dense breasts, with higher contribution in women younger than 50 years\cite{Corsetti2008}. Moreover, breast ultrasound is simple, effective and low cost. Because of all these advantages, it can be applied in a large scale for imaging, for example, in China\cite{Song2015}. 

\begin{figure}[ht]
\centering
\includegraphics[scale=0.4]{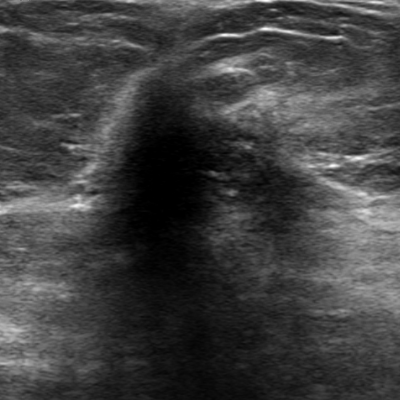}
\caption{A malignant lesion in breast ultrasound}
\label{fig:example}
\end{figure}

In the clinical workflow of breast ultrasound imaging, radiologists often report the sizes of breast lesions, describe the lesions according to 
BI-RADS lexicon\cite{DOrsi2013} and estimate the final BI-RADS score. An accurate delineation of breast lesion can help radiologists to describe margin, shape and posterior features. However, manual segmentation of breast lesions is time-consuming and tedious. 
The segmentation also varies from one reader to another. Therefore, the automated segmentation can play a key role in facilitating the reporting of the diagnosis. In terms of detection and diagnosis, 
computers can also assist radiologists to make decisions that improve the effectiveness of ultrasound reading. For example, computer techniques{\cite{Kozegar2017,Tan2015,Tan2013,Tan2012,Tan2013a,Liu2014,raha2017fully,fleury2018the}}
have been proposed to delineate the contour of lesions or directly detect or diagnose breast lesions. 
% * <byw94@hotmail.com> 2018-11-23T16:00:17.370Z:
% 
% have been proposed to delineate the contour of lesions, directly detect and diagnose breast lesions.
% 
% ^.
Most of these computer-aided diagnoses or detections include a module of segmentation. Therefore, it is important to develop a robust and accurate segmentation method.

Breast lesion segmentation is very challenging, especially when there is the presence of noise, the ill-defined edges, irregular shapes, and different posterior behaviors of
lesions. As Fig. \ref{fig:example} shows, there is strong shadowing in the posterior and upper region, the lesion boundary is fuzzy and not clear. Therefore, there is a risk that segmentation algorithms fail, 
causing oversegmentation. 

There are two types of segmentation methods: contour-based and region-based methods. The contour-based segmentation relies on finding the optimal contour to enclose the whole breast lesion. Region-based methods aim to assign a label to every image pixel. Jing et al. \cite{Cui2009} proposed an iterative segmentation scheme to refine the initial contour and perform self-examination and correction on the segmentation result. Their best intersection of the computer and the reference segmented area was 0.84. Tan et al. \cite{Tan2016b} proposed a novel depth-dependent dynamic programming technique and obtained a Dice similarity coefficient (DSC) of 0.73. This accuracy was then improved by Kozegar et al. with a specific level set algorithm\cite{Kozegar2018}. Horsch et al.\cite{Horsch2001} presented a computationally efficient segmentation algorithm for breast masses on sonography, which is based on maximizing a utility function over partition margins defined through gray value thresholding of a preprocessed image. Their algorithm was evaluated on a database of 400 cases and the reported average overlap rate was 0.73. The challenge of applying contour-based method is to make sure the contour evolvement is not trapped by non-breast edges. For region-based segmentation, both traditional methods and machine-learning-based pixel classification methods were investigated. Feng et al. \cite{feng2017an} adopted an adaptive fuzzy C-means algorithm and the obtained DSC is 0.925. Pons et al. \cite{Pons2016a} reported that their evaluated automated method achieved a DSC of 0.49 using a Markov Random Field (MRF) and a Maximum a Posteriori (MAP) approach, by applying it to clinical data. Agarwal et al.\cite{Agarwal2018} developed a semi-automatic framework for breast lesion segmentation in ABUS volumes which is based on the Watershed algorithm. Rodrigues et al. \cite{Rodrigues2015} took the advantage of pixel-wise classification and achieved a DSC of 0.824. Kumar et al.\cite{Kumar2018a} proposed convolutional neural network approaches for breast ultrasound lesion segmentation and their algorithms effectively segmented the breast masses, achieving a mean DSC of 0.82. 
% * <lizheren0613@163.com> 2018-12-11T13:06:19.069Z:
% 
% > Their best intersection of the computer and the reference segmented areas to the reference segmented area was 0.84. 
% Their best intersection of the computer and the reference segmented areas  was 0.84. 
% 
% ^.

As one branch of machine learning, deep learning has become popular as a self-taught approach in which features are computed in an automatic manner instead of combining manually designed features\cite{Krizhevsky2017,He2016Identity,ioffe2015batch,huang2016densely,Tan2018,Jiang2018b}. These approaches have rapidly become state-of-the-art that outperform other traditional methods in the segmentation tasks with ultrasound. There are generally two ways of applying deep learning: patch-wise classification using convolutional neural networks (CNN) and pixel-wise classification using fully convolutional networks (FCN) such as ResNet and U-Net architecture\cite{Ronneberger2015}. These techniques have gained 
propitiatory for the segmentation tasks. Most existing deep learning based methods still rely on image information (lesion boundaries) while the prior knowledge of breast lesion shape is not well used, although they have already obtained accurate segmentation results. To further improve classification results, proper incorporation of prior knowledge is necessary. In this work we use a model which tends to learn prior knowledge of breast lesions and is able to properly deal with fuzziness or even the absence of a visible lesion edge in some parts of lesions. We proposed a generative adversarial net (GAN) based framework, semi-pixel-wise cycle generative adversarial net (SPCGAN), for segmenting the lesion in 2D ultrasound images. 

The main contributions of this work are as follows: 1. We propose a new breast lesion segmentation method that uses a deep learning approach where we combine  the (cycle) GAN loss with a FCN loss (to also include a pixelwise classification). 2. We show clear improvements over existing lesion segmentation approaches on 2D ultrasound image datasets  of breasts. The combination will make the segmentation not only reply on lesion boundary but also mimic the way of human annotation. Because of the power of GAN, the scheme requires less data to train a model with effectiveness and robustness.

\begin{figure*}[ht]
\centering
\includegraphics[scale=0.15]{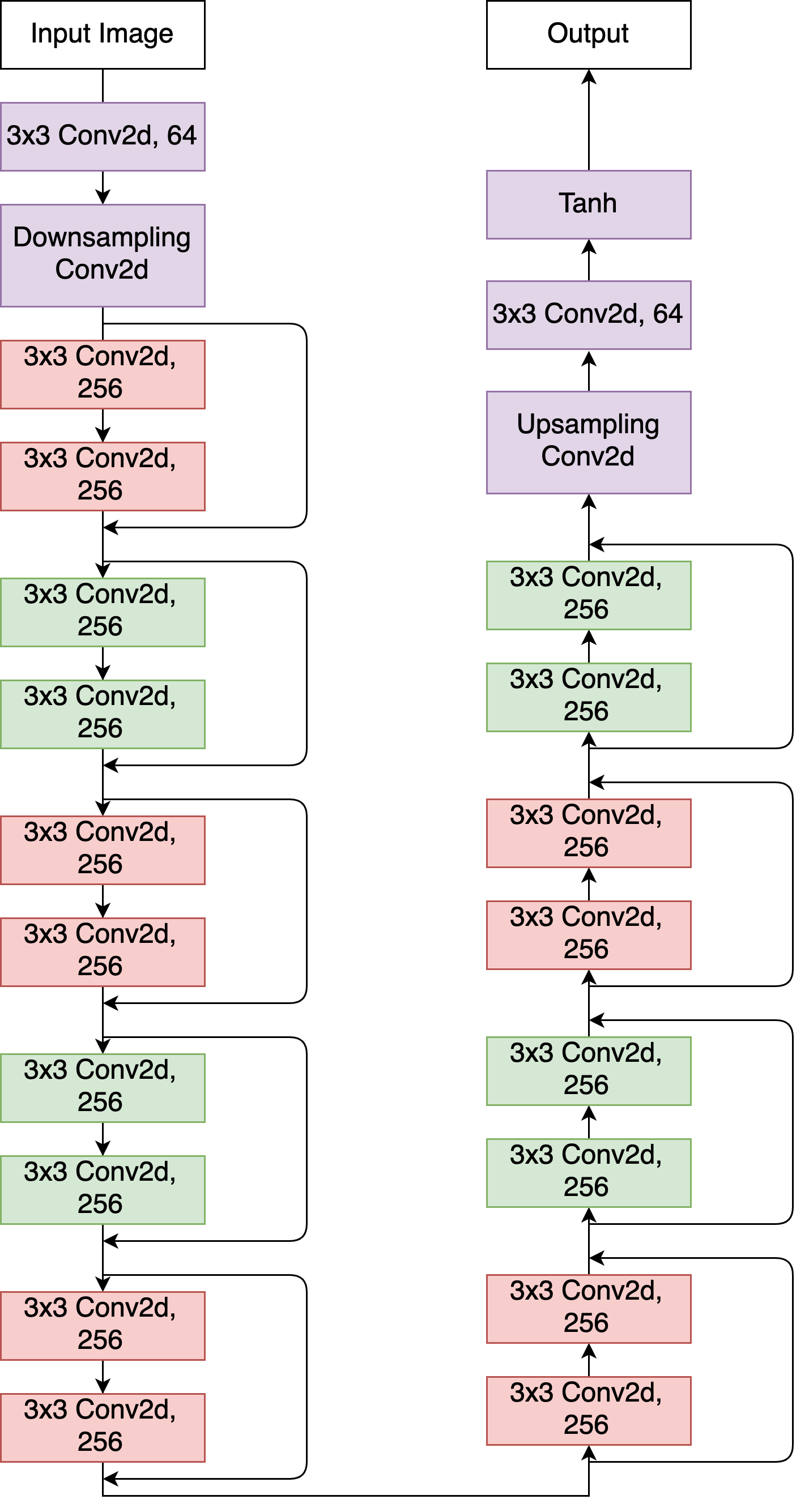}
\caption{The visualization of the structure of full convolutional networks, ResNet. The networks contains 9 residual blocks. Each block consists of two convolutional layers. Downsampling is performed on the input image before it was forward to the residual blocks and upsampled after it finished all the residual calculations. The size of input and output images are the same.}
\label{resnet}
\end{figure*}

\section{Methods}

\subsection{Semi-pixel-wise Cycle Generative Adversarial Net}

%introduce GAN and cycle GAN, introduce the loss of the cycle gan

%introduce the added loss from the generator
{Generative adversarial net (GAN) is a framework which consists of two models for the estimation of generative results via an adversarial process} \cite{goodfellow2014generative}. There is a generative model $G: \mathbb{R}^N \to C(\mathcal{N}, \mathbb{L}_{1}(\Omega))$ given by $\mathbf{w} \mapsto G_{\mathbf{w}}$ which tries to produce data that is similarly distributed as training data. G
Here the set of noisy images denoted by $\mathcal{N}$ is obtained by sampling of uncorrelated Gaussian processes per position $\mathbf{x} \in \Omega$ in the rectangular image domain $\Omega \subset \mathbb{R}^2$ on which all images are supported.
Then given variable weights $\textbf{w}\in \mathbb{R}^N$ it constructs a continuous mapping $G_{\textbf{w}}$ from the set of noise images $\mathcal{N}$ to a new set. Henceforth, this new set is called the set of fake images $\mathcal{F}:=G_{\mathbf{w}}(\mathcal{N}) \subset \mathbb{L}_{1}(\Omega)$. 
We also have a set of true images $\mathcal{T} \subset{L}_{1}(\Omega)$. In the training, we assign true images with label $1$.

Simultaneously, a discriminative model
$\mathbb{R}^M \ni \textbf{v} \mapsto D_{\textbf{v}}$ where operator
$D_{\textbf{v}}: \mathbb{L}_{1}(\Omega) \to \{0,1\}$ evaluates the authenticity of a sample data coming from training set, and where $\mathbf{v}$ are weights of the discriminative model. 
We apply a loss function to the pair of operators/models $D$ and $G$ to discriminate whether the input sample is real or not.
Here both G and D aim to deteriorate the performance of each other, therefore the loss function of GANs mimics a two-player mini-max game and is expressed as follows:
\begin{equation} \label{game}
\begin{array}{l}
\min\limits_{\textbf{w} \in \mathbb{R}^N} \max\limits_{\textbf{v} \in \mathbb{R}^M} \mathcal{L}_{GAN}(D_{\mathbf{v}},G_{\mathbf{w}},\mathcal{T},\mathcal{N}),  \\
\textrm{with the loss-function given by } \\[6pt]
\mathcal{L}_{GAN}(D_{\mathbf{v}},G_{\mathbf{w}},\mathcal{T},\mathcal{N}) \\[6pt]
=\mathbb{E}_{p_{data}}[\log D_{\boldsymbol{v}}(\cdot)]+\mathbb{E}_{p}[\log (1-D(G_{\boldsymbol{w}}(\cdot))] \\[6pt]
=\frac{1}{|\mathcal{T}|} \sum \limits_{f \in \mathcal{T}} \log (D_{\mathbf{v}}f) +
\frac{1}{|\mathcal{N}|} \sum \limits_{z \in \mathcal{N}} \log (1-D_{\mathbf{v}}(G_{\mathbf{w}}z) ). 
\end{array}
\end{equation}
where $\mathbf{w}$ are the weights for the generator model G and $\mathbf{v}$ are the weights of the discriminator model D.
%where $p_{\boldsymbol{z}}(\boldsymbol{z})$ is the distribution of training data and $G(\boldsymbol{z})$ is the generative distribution over training data $\boldsymbol{x}$. 
The map $D_{\boldsymbol{v}}f$ gives the probability that data $f \in \mathbb{L}_{1}(\Omega)$ is a real image rather than a fake one (meaning there does not exist a noise image $f_0 \in \mathcal{N}$ such that $f=G_{\mathbf{w}}f_0$). Note that $p_{DATA}(f)=\frac{1}{|\mathcal{T}|}$ is a uniform distribution over the true images and $p$ is a uniform distribution over the noisy images $p(z)=\frac{1}{|\mathcal{N}|}$ where $|\mathcal{N}|$ denotes the total number of noisy images (which is also the total number of fake images if %we assume 
$G_{\mathbf{w}}$ is injective).

Note that in (\ref{game}) the distriminative model $D$ is trained to maximize the probability to assign correct labels to input data $f \in \mathbb{L}_{1}(\Omega)$. Meanwhile, the generative model $G$ is trained to disturb the judgment by the discriminative model $D$. 
Since the first term in (\ref{game}) is independent of $G$ this is done by minimizing the second term which is the expectation \mbox{$\mathbb{E}_{p}[\log (1-D_{\mathbf{v}}(G_{\boldsymbol{w}}(\cdot)))]$}. 
% * <farhad.ghazvinian@gmail.com> 2018-12-14T07:51:01.147Z:
% 
% >  
% somewhere the variable z needs to be introduced
% 
% ^.
% * <farhad.ghazvinian@gmail.com> 2018-12-14T07:46:19.316Z:
% 
% > is the distribution of training data
% perhaps it is the distribution of the latent variables!
% 
% ^.
%% insert the actual application of the adversarial net in our model and specify the corresponding parameters in our dataset.

In a CycleGAN model\cite{CycleGAN2017}, the generator no longer generates data from random source images $\mathcal{N}$ such as white noise. There are two target domains $\mathcal{T}_1, \mathcal{T}_{2} \subset \mathbb{L}_{1}(\Omega)$, which are sets of true images and which can be unpaired for data transfer between each other and the data generation process is now drawn in analogy to an autoencoder. There are two generators to translate data in one domain to the other, which can be regarded as an encoder and decoder respectively. There are also two discriminators and each of them tries to discriminate the authenticity of the data that belongs to the corresponding domain. Assume we have two real images $T_1, T_2 \in \mathbb{L}_{1}(\Omega)$ with $T_{1} \in \mathcal{T}_1$ and $T_{2} \in \mathcal{T}_{2}$, then the design of the CycleGAN model is as follows:
\begin{equation}
\begin{aligned}
\{0,1\}\stackrel{D_{\mathbf{v}_1}}{\longleftarrow} \underset{F_1}{T_1}  \xtofrom[G_{\mathbf{w}_2}]{G_{\mathbf{w}_1}} \underset{T_2}{F_2}  \stackrel{D_{\mathbf{v}_2}}{\longrightarrow}\{0,1\},
%\rightleftharpoons[G_{w_2}]{G_{w_1}}
\end{aligned}
\end{equation}
where $D_{v_k}(T_k)=1$ if $T_{k} \in \mathcal{T}_{k}$ and $0$ else, for $k=1,2$,
and where fake images are given by $F_{2}=G_{\mathbf{w}_1}T_1$ and $F_{1}=G_{\mathbf{w}_2}T_2$.
\begin{comment} % I would say this figure only gives confusion G_A2B with two for to
% The A and B where we have \mathcal{T}_{1} and \mathcal{T}_{2}...Fake A is output of the 
% operators which is not immediately clear etc. etc.
% It only shows the first term in (3) which shows an asymmetry in th figure etc. etc. 
The complete generation and discrimination process forms a cycle as shown in Figure \ref{cycgans}.
\begin{figure*}[ht]
\centering
\includegraphics[scale=0.14]{image/cycgans.jpg}
\caption{The architecture of CycleGAN model: forward GAN on the upper half is aiming to transform an image from Domain $A:=\mathcal{T}_{1}$ to Domain $B=\mathcal{T}_{2}$ via generator $G_{AB}$ and being distinguished by discriminator $D_{B}$ from Domain $B$. Backward GAN on the bottom half is working in the opposite way, it take an image in Domain $B$ and transforms it into Domain $A$. Then $D_A$ distinguishes the generated image in Domain A from the real one.}
\label{cycgans}
\end{figure*}
\end{comment}

\begin{figure*}[ht]
\centering
\includegraphics[scale=0.15]{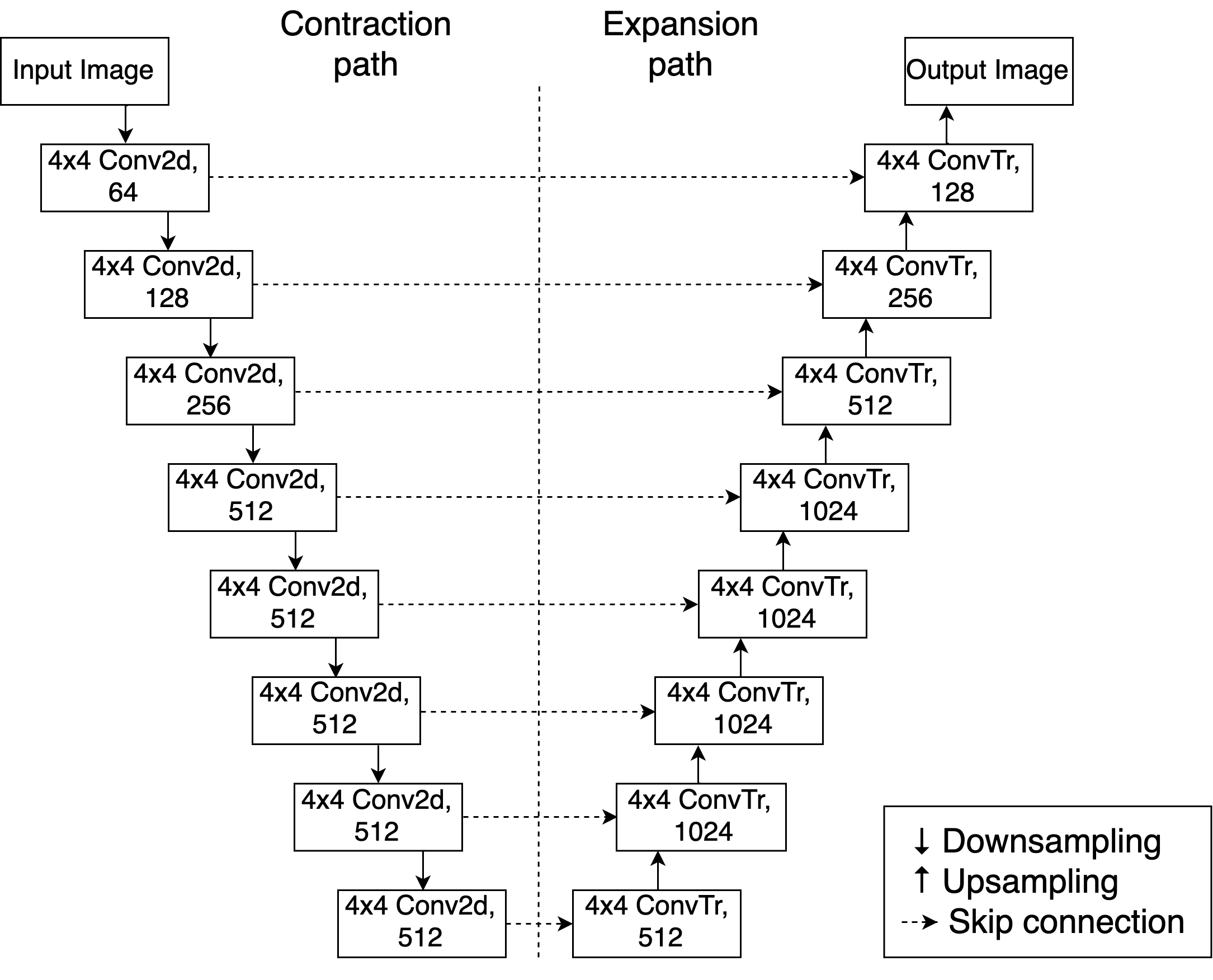}
\caption{The general architecture of the generators $G_{AB}$ and $G_{BA}$ in our model. It is an FCN based model and has the structure of U-Net. On the left side is the downsampling process which extracts the feature maps via convolutional layers from input image. On the right side is upsampling via convolutional transpose layers and connects with the feature map with upsampling result to get the final output.}
\label{unet}
\end{figure*}

Comparing to transfer data between target domains via two GANs, the cycle mechanism of CycleGAN network necessarily guarantees the one-to-one mapping relationship between the input and output data and therefore rules out the possibility that any input data can be mapped to induce a set of output data distributions which match the target domain\cite{CycleGAN2017}. 

Let us include an $AB$-labeling in $G_{AB}$ to stress that it is a generative model from `domain' $A$ to 
`domain' $B$ which typically differs from the generative model $G_{BA}$ from $B$ to $A$. %QUESTION: HOW DOES IT PRECISELY DIFFER IN THE APPLICATION ??
With domain we mean the joint set of the set of true images and fake images: 
%QUESTION: DO we assume these sets are disjoint ???   % Then  %PLEASE CHECK THE FOLLOWING:
\begin{equation} \label{AB}
A=\mathcal{T}_{1} \cup \mathcal{F}_{1} 
\textrm{ and }B=\mathcal{T}_{2} \cup \mathcal{F}_{2}, \textrm{ with \ }\mathcal{T}_i \cap \mathcal{F}_i =\emptyset, 
\end{equation}
where the sets $\mathcal{F}_1, \mathcal{F}_{2}$ of fake images are obtained by letting the generator act on the true-images of the other domain, i.e. 
\begin{equation} \label{fake}
\mathcal{F}_{1}= G_{BA,\mathbf{w}_2}(\mathcal{T}_{2}) \textrm{ and  }\mathcal{F}_2=G_{AB,\mathbf{w}_1}(\mathcal{T}_{1}).
\end{equation}
%\begin{remark} 
\emph{Remark}. %(domains in our application setting) \\
In our application of segmenting lesions in ultrasound images, $\mathcal{T}_{1}$ is the set of ultrasound images without segmentation, and $\mathcal{T}_2$ is the set
of manually segmented images. Then our generative convolutional network models are according to the design in Fig. \ref{unet} as we explain below. 
Then $A$ and $B$ are given by (\ref{AB}) and (\ref{fake}).
%\end{remark}
\ \\ \\
 To ensure the expected mapping from input to the desired output, there is also a cycle loss to evaluate the decoder performance, which is to check whether the transformed data can be brought back to the original domain in the generative model transformation % not `translation'
 from domain $B$ to domain $A$, and vice versa.
 This then provides cycle consistency in both forward and backward direction. Summarizing, we need
 \[
 \begin{array}{ll}
 z \rightarrow  G_{BA, \mathbf{w}_2}(G_{AB,\mathbf{w}_1}(z))\approx z \textrm{ for all }z \in \mathcal{T}_1, \\
 z \rightarrow  G_{AB, \mathbf{w}_1}(G_{BA,\mathbf{w}_2}(z))\approx z \textrm{ for all }z \in \mathcal{T}_2.
  \end{array}
 \]
To achieve this the cyclic loss function is expressed as follows:
\begin{equation}
\begin{array}{l}
\mathcal{L}_{cyc}(\mathbf{w}_1,\mathbf{w}_2,\mathcal{T}_1,\mathcal{T}_2)= \\
 \; \,
\mathbb{E}_{p_{dataA}}\left[\; \|G_{BA,\mathbf{w}_2}(G_{AB,\mathbf{w}_1 }))(\cdot)-(\cdot) \|_{\mathbb{L}_{1}(\Omega)} \right]\\
+ \mathbb{E}_{p_{dataB}} \left[ \|G_{AB,\mathbf{w}_1}(G_{BA,\mathbf{w}_2 }))(\cdot)-(\cdot) \|_{\mathbb{L}_{1}(\Omega)} \right]\\
\\
=\frac{1}{|T_1|}\sum \limits_{f\in \mathcal{T}_1}\int \limits_{\Omega} |\; [G_{BA, \mathbf{w}_2}(G_{AB, \mathbf{w}_1}(f)](\mathbf{x})-f(\mathbf{x})\;| \, {\rm d}\mathbf{x}\\
+\frac{1}{|T_2|}\sum \limits_{f\in \mathcal{T}_2}\int \limits_{\Omega} |\;[G_{AB, \mathbf{w}_1}(G_{BA,\mathbf{w}_2}(f)]\mathbf{x})-f(\mathbf{x})\;| \, {\rm d}\mathbf{x}.
% f here is confusing, TO BE dISCUSSED
\end{array}
\end{equation}
%ABOVE UPDATED WITH TAN AND REMCO...NEXT TIME WE CONTINUE HERE......
\begin{comment}
where $w_1$ is the input image to be transferred to domain B and $w_2$ is the input image to be transferred to domain A. $T_1$ and $T_2$ are the transferred images in domain A and B respectively.
\end{comment}

Eventually, in the CycleGAN model, the \emph{ total loss }$\mathcal{L}$ is expressed as follows:
\begin{equation}
\begin{aligned}
&\mathcal{L}((\mathbf{w}_1,\mathbf{v}_1,\mathcal{T}_1),(\mathbf{w}_2,\mathbf{v}_2,\mathcal{T}_2))=\\
&\mathcal{L}_{GAN}(D_{\mathbf{v}_{1}},G_{\mathbf{w}_1}, \mathcal{T}_1,\mathcal{T}_2)+\mathcal{L}_{GAN}(D_{\mathbf{v}_{2}},G_{\mathbf{w}_2}, \mathcal{T}_2,\mathcal{T}_1)\\
&+\mathcal{L}_{cyc}(\mathbf{w}_1,\mathbf{w}_2,\mathcal{T}_1,\mathcal{T}_2)
\end{aligned}
\end{equation}
Now suppose optimization (\ref{game}) is performed over the training dataset. This gives optimum $(\mathbf{w}_1^*,\mathbf{w}_2^*,
\mathbf{v}_1^*,\mathbf{v}_2^*)$ that we hope to find efficiently by stochastic gradient descent in the usual deep learning approach.

Then given
a test image $f \in \mathbb{L}_{1}(\Omega)$ the
output segmentation is %given by
\begin{equation}
\Omega \ni \mathbf{x} \mapsto (G_{AB,\mathbf{w}_1^*}f)(\mathbf{x}) \in \mathbb{R},
\end{equation}
where the output values are typically close to $\{0,1\}$ due to the setting of manually segmented images in $\mathcal{T}_{2}$ in the training set.

In this work we adopt the general architecture of CycleGAN to our model and manipulate on the discriminator loss part by adding an extra loss related to the pixel-wise classification. %QUESTION: MORE CLEAR FORMULATION NEEDED. PLEASE CHECK BELOW
This means that 
\begin{equation} \label{Lnew}
\begin{array}{l}
\mathcal{L}^{NEW}((\mathbf{w}_1,\mathbf{v}_1,\mathcal{T}_1),(\mathbf{w}_2,\mathbf{v}_2,\mathcal{T}_2)) \\ =
\mathcal{L}((\mathbf{w}_1,\mathbf{v}_1,\mathcal{T}_1),(\mathbf{w}_2,\mathbf{v}_2,\mathcal{T}_2))
+
\mathcal{L}_{pix-wise}(\mathbf{w}_1,\mathcal{T}_1,\mathcal{T}_{2}).
\end{array}
\end{equation}
Now we have a correspondence between raw training images and their corresponding segmentation. Let us therefore write $f_{1} \sim f_{2}$ if raw image $f_{1} \in \mathcal{T}_1$ and manually segmented image $f_{2} \in \mathcal{T}_2$ is indeed the segmentation of image $f_1$. Note that
$|\mathcal{T}_1|=|\mathcal{T}_2|$. Then we set
\begin{equation} \label{PWLOSS}
\begin{array}{l}
\mathcal{L}_{pix-wise}(\mathbf{w}_1,\mathcal{T}_1,\mathcal{T}_{2})= 
\frac{1}{|\mathcal{T}_1|}\! \!\!\sum \limits_{
{\tiny
\begin{array}{c}
f_1 \in \mathcal{T}_{1}, \\ f_2 \in \mathcal{T}_{2}, \\ f_1 \sim f_2
\end{array}
}} \!\!\!\! MSE(G_{BA,\mathbf{w}_1}(f_1),f_2),  
\end{array}
%\label{pixelloss}
\end{equation}
where the Mean Square Error (MSE) is given by
$MSE(f,g)=\frac{1}{|\Omega|} \int_{\Omega} |f(\mathbf{x})-g(\mathbf{x})|^2 \, {\rm d}\mathbf{x}
$.\\
% * <farhad.ghazvinian@gmail.com> 2018-12-14T08:09:25.286Z:
% 
% > The probability distribution of $z$ in the latent space is then aligned to the real data distribution. 
% I do not think so! The distribution of z does not change. The network learns to fold this distribution to match as much as possible to the distribution of the target (data).
% √√√modified
% ^.

\begin{figure*}[ht]
\centering
\includegraphics[scale=0.6]{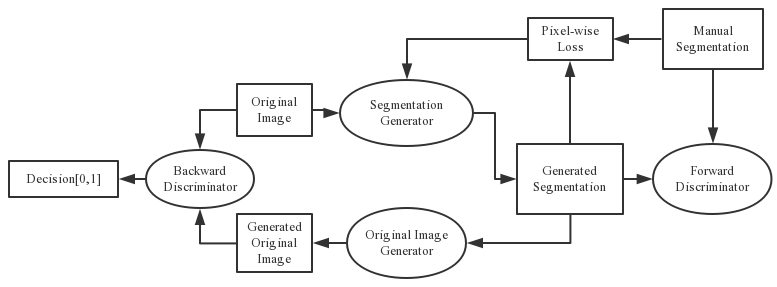}
\caption{The architecture of SPCGAN with pixel-wise loss. The pixel-wise loss is only applied in the forward cycle. In the forward cycle, the segmentation generator receives unsegmented original image and produces an auto-segmented image. The forward discriminator receives both auto-segmented and ground truth segmentation images to make pixel-wise classification. In the backward cycle, the auto-segmented image is fed into the backward generator to obtain a cyclic original image and input to the backward discrinator along with the real original one.}
\label{pxcycgans}
\end{figure*}

Now, we consider that $\mathcal{T}_1$ is the set of images without segmentation. $\mathcal{T}_2$ is the set of images with corresponding manual annotations.
By adapting the loss function (\ref{PWLOSS}) the optimal weights $(\mathbf{v}_{1}^*,\mathbf{v}_{2}^*)$ of the discriminative part are changed in such a way that discriminator stay close to the pixel-wise correct classifications for each pair $(f_1,f_2)$ of ultrasound image $f_1 \in \mathcal{T}_{1}$ and its manual segmentation $f_2 \in \mathcal{T}_2$.
This evaluates the adversarial loss between the manually annotated segmentation and the generated segmentation in every pixel (\ref{PWLOSS}): \\ 
Rather than just considering the probability of classifying the whole image 
%This pixel-wise loss is calculated by the following formula:
%
% \begin{equation}
% \begin{aligned}
% \mathcal{L}_{\text{pixel-wiseGAN}}(G,D)&=\sum^{N}_{i=1}{\frac{\|\boldsymbol{G_i}-\boldsymbol{D_i}\|^2_2}{N}},
% \end{aligned}
% \end{equation}
in the forward process of the cycle, the discriminator tries to defy and reject the generated (pixel-wise) segmentation. Now the generator will minimize the pixel-wise loss 
so that the generated segmentations will be accepted by the discriminator. The architecture of our model is shown in Fig.~\ref{pxcycgans}.
% * <farhad.ghazvinian@gmail.com> 2018-12-14T08:04:54.271Z:
% 
% > For each pixel pair between the two segmentations, we calculate the difference to obtain the pixel-wise loss and sum up over the whole image then divided by the total number of pixels N to obtain the average.
% 
% Can easily be replaced by "mean square error (MSE)" as it is well-known. 
% √√√modified
% ^.
% * <farhad.ghazvinian@gmail.com> 2018-12-14T08:01:33.927Z:
% 
% > where G is the generated segmentation image and D is the ground truth segmentation.
% These letters (G and D) already have been used as the network's function. Other choices are preferred for avoiding possible confusion.
% √√√modified
% ^.
In the GAN-algorithm, the input data is drawn from a simple prior probability distribution such as a Gaussian distribution. Thereby, the input is essentially a latent vector of unstructured noise. The network learns to fold the probability distribution of the input data in the latent space to match as much as possible to the distribution of the target data. It is not able to take the advantage of the prior knowledge of images. However, in a CycleGAN-like algorithm, because there are two domains, we could draw the prior knowledge from the source image as the prior distribution and then generate samples based on it rather than imposing a random probability. In our implementation, the prior knowledge of annotated segmentation of breast lesions is learned by the model so that it is able to properly deal with images with ambiguous features, for example, in the absence of visible lesion edges. 

\subsection{Fully Convolutional Network (FCN) Generator }
The generator used above itself is an FCN and this type of network is applied extensively in ultrasound \cite{Zhang2016a}\cite{Yap2017}\cite{Ravishankar2017a ,Wu2017,Oktay2018,Milletari2017} \cite{Sundaresan2017a}\cite{Wu2017b} for various applications.

% To show the generic benefits of our model we tested with 2 different deep learning architectures (that is ResNet and U-Net) in the FCN part (given by Eq. \ref{PWLOSS}) of our model.

To show the advantage of our model, we  applied two networks for the FCN structure {(given by Eq. \ref{PWLOSS})}, U-Net {shown in Fig. \ref{unet} and an FCN based on ResNet shown in Fig.\ref{resnet}. Both model types integrate the features of FCN so that they can be trained end-to-end by upsampling and deconvoluting the feature maps extracted from convolutional layers and finally output pixel-wise classifications on the input images. It is widely used in semantic segmentations\cite{long2015fully}.} We compared the performances of our framework combined with these two structure models respectively. Throughout this manuscript we refer to U-Net and ResNet as the "backbone" of the FCN. This is done to indicate the deep-learning architecture that underlies the FCN part of our model.

% TODO: modify figure for UNet

U-Net is a type of improved FCN. It can be regarded as two parts: the first half is in charge of the feature extraction as the convolutional layers work in CNN. The second half is to upsample the extracted feature which allows the context information to be propagated to higher resolution layers and output the segmented image as the same size as input. The U-Net structure we constructed in this experiment is shown in Fig. \ref{unet}. 

ResNet\cite{DBLP:journals/corr/HeZRS15} aims to optimize the deteriorated performance of networks with very deep layers. It uses a residual block to create a shortcut connection in order to skip some of the convolutional layers. In this type of network, we could reduce the number of parameters so that the computation is simplified. The plain network architecture of ResNet keeps hierarchical configuration. The number of feature maps increase along with depth and the ability of feature extraction is therefore guaranteed. In our experiment, we applied a ResNet with 9 residual blocks as shown in Fig. \ref{resnet} but modified to an FCN-like model which contains downsampling and upsampling before and after residual calculation respectively. This modification makes our ResNet model adapt to the segmentation task rather than the classification as the original one does.

\section{Compared Methods}
\subsection{The Level Set Method}

{To compare our deep learning scheme with a traditional segmentation method, we applied a geodesic-active-contour (GAC) based level set method. This level set method includes curvature and advection terms as introduced by Caselles et al. \cite{Case97}. The partial differential equation describing the motion of the contour is defined by: }

\begin{equation}
\frac{\partial }{\partial t}\Phi +g\cdot(1+\varepsilon k)|\nabla\Phi|+\alpha \nabla g\cdot \nabla \Phi =0
\end{equation}

\noindent with $(\mathbf{x},t) \to \Phi(\mathbf{x},t)$ the level set function, $\varepsilon$ the curvature influence, $k$ the curvature of the level set, $\alpha$ the advection influence and $g$ the image gradient-based speed function given by
\begin{equation}
 g(\mathbf{x}) = 1 / ( 1 + | (\nabla G_s *f)(\mathbf{x})| ) ,
\end{equation}
\noindent where $f$ is the input test image and 
$(\nabla G_s *f)(\mathbf{x})$ is the derivative of Gaussian operator applied at scale of $s=\frac{1}{2}\sigma^2$ with a standard-deviation that we manually set to $\sigma= ...\textrm{pixels}$. Note that the segmentation boundary is given by the 0-level set \mbox{$\{\mathbf{x} \in \Omega \;|\; \Phi(x,t)=0\}$} where both the evolution time $t>0$ and the parameters $\varepsilon,\alpha >0$ 
are optimized via a gradient search algorithm applied on the training set, using the average Dice coefficient (\ref{DSC}) criterium.

 We initialize the level set with the center of the lesion. The segmentation can be controlled by setting the weights ($\alpha$, $\varepsilon$) of propagation, curvature and advection term. The propagation term controls the inflation or `balloon force' of the segmentation, the curvature term controls the `smoothness' of the boundary of the segmentation and the advection term in the update equation  attracts the contour to the lesion edge\cite{Tan2016b}. 

\subsection{Mask R-CNN Method}
{Mask R-CNN model has excellent performance in object detection and instance segmentation. For the purpose of comparison, we also implemented this model as a reference to evaluate our proposed model. Mask R-CNN model extends Faster R-CNN by adding a branch to predict an object mask along with the existing branch of bounding box recognition\cite{DBLP:journals/corr/HeGDG17}. Faster R-CNN outputs bounding boxes and the classification and bounding-box regression, while Mask R-CNN adopts both of the outputs from Faster R-CNN as well as a binary mask for each candidate object. Due to the modification on region of interest (RoI) Pooling layers in Faster R-CNN to RoI Align layers, the misalignment issue has been solved and therefore, prediction with higher accuracy in spatial layout can be achieved, which is good for detecting the edge of tumors in breast cancer images.}

% \textbf{add reference of cycle} 
% \textbf{add architecture in figure of G and D of cycle GAN we used}
% \textbf{explain why GAN can learn prior knowedlge, read this https://arxiv.org/abs/1806.03577}

\section{Materials}

\subsection{Datasets}
This study is based on 2D BUS DICOM images of abnormal patients. All DICOM images were scanned from SIEMENS MED SMS USG S2000 and TOSHIBA Aplio400 TUS-A400 Ultrasound System. For this study, we collected a dataset of 670 breast lesion ultrasound images from different women (aged 18-70) that had no history of breast cancer. Among the 670 images, 640 were scanned by a SIEMENS Ultrasound System and 30 were scanned by a TOSHIBA Ultrasound System. If there were a number of DICOM images from the same lesion, the DICOM image containing the maximum area of the lesion was collected into the dataset. The type of lesion has been clinically diagnosed as malignant or benign. Among the 640 lesions from SIEMENS Ultrasound System, 120 are malignant lesions and 520 are benign lesions. 

% * <lizheren0613@163.com> 2018-12-12T06:20:24.970Z:
% 
% And the 30 lesions from TOSHIBA Ultrasound System are malignant lesions.
% 
% ^.

\begin{figure}[ht]
\centering
\includegraphics[scale=0.6]{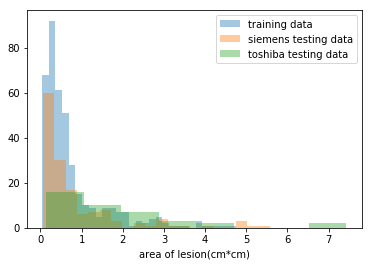}
\caption{The histogram of lesion areas in training and testing dataset.}
\label{fig:distplot}
\end{figure}

In our study dataset, for SIEMENS images, 399 were used in the training phase, 141 were used in the testing phase and 100 were used in the validation phase. All TOSHIBA images were used for testing the generality of the model trained by SIEMENS images. During the image preprocessing stage, the original DICOM images were re-sampled to 0.1mm spacing in both horizontal and vertical directions. After that the ROI images were cropped from the re-sampled DICOM images with a size of 400*400 for easy processing, and the center of the lesions were the center of the ROI images.

For each ROI image, we manually generated reference lesion segmentations by using a MATLAB program. These manual segmentations were performed by an experienced researcher with 10 years of experience in breast ultrasound. 

\subsection{Performance Evaluations}
% * <farhad.ghazvinian@gmail.com> 2018-12-14T08:19:03.100Z:
% 
% > Performance Evaluations
% maybe measuring additional metrics like average precision and recall show a comprehensive evaluation 
% 
% ^.
In this study, Dice similarity coefficient is used for describing the accuracy of the segmentation by different methods. Dice similarity coefficient is a statistic used for comparing the similarity between two samples, and is defined as follows:

\begin{equation} \label{DSC}
\emph{DSC} = \frac{2\left|X \cap Y \right|}{\left|X\right|+\left|Y\right|},
\end{equation}

where $\left|X\right|$ is the area of lesion by manual segmentation and $\left|Y\right|$ is the area of lesion segmented by automatic methods. The larger the Dice similarity coefficient is, the higher accuracy of the computational segmentation is. It ranges between 0 and 1.

\subsection{Implementation platform}
Our network was trained on a workstation equipped with a NVIDIA GeForce GTX 1080Ti GPU {and an Intel i7-7700 CPU}. Our implementation is built on the top of CycleGAN release \footnote{https://github.com/junyanz/pytorch-CycleGAN-and-pix2pix.} 
% * <farhad.ghazvinian@gmail.com> 2018-12-14T08:23:17.849Z:
% 
% > Our network was trained on a workstation equipped with an NVIDIA GeForce GTX 1080Ti GPU
% This is relevant if the paper reports the processing/execution time. Perhaps it is worth to mention early that the adversarial setting does not add any extra processing time in the execution time (prediction mode) as it is only involved in training.
% 
% ^.

\section{Experiments and Results}
To show the superiority of our framework, we test SPCGAN, the backbone structure (ResNet) only and the level set method on both Siemens and Toshiba dataset. We also test the effectiveness of only applying generative adversarial networks framework with pixel-wise cost (GAN) without the cycle loss. To indicate the applicability of this framework, we also compare different frameworks but with a different backbone FCN structure (U-Net). For all experiments we performed training with the number of epochs 1500 with mini batch size 1, Adam optimizer \cite{kingma2015adam:} and without dropout. {We adopt training process from Zhu et al.\cite{CycleGAN2017}. We first train our networks from scratch with a
learning rate of 0:0002. Then we keep the same learning rate for the first 750 epochs and linearly decay
the rate to zero over the next 750 epochs. Weights are initialized from a Gaussian distribution N(0;
0:02).}  We use validation set to choose the model with least loss from different epochs. 

We also apply the following data augmentation methods: shear (0.2 range), rotation (10 range), width shift (0.1 range), height shift (0.1 range), zoom (0.1 range), and horizontal flip to solve the problem of insufficient data.

\subsection{Statistical Analysis}
%修改
One-sided paired t-tests are used for statistical analysis when comparing results from different segmentation methods.
The hypothesis in this study will be tested to control type I error rate at alpha = 0.05. The hypothesis is that the DSC of the SPCGAN is superior to (bigger than) that of ResNet or the level set method, for statistical significance level alpha = 0.05.

%修改
%cyclegnn,ResNet:statistic=4.191451,pvalue=4.885899e-05

%cyclegnn,levelset:statistic=10.890803, pvalue=2.176008e-20

%ResNet,levelset:statistic=8.334157, pvalue=6.515888e-14

\begin{figure}[htbp]
\centering
\subfigure[]{\includegraphics[width = .165\linewidth]{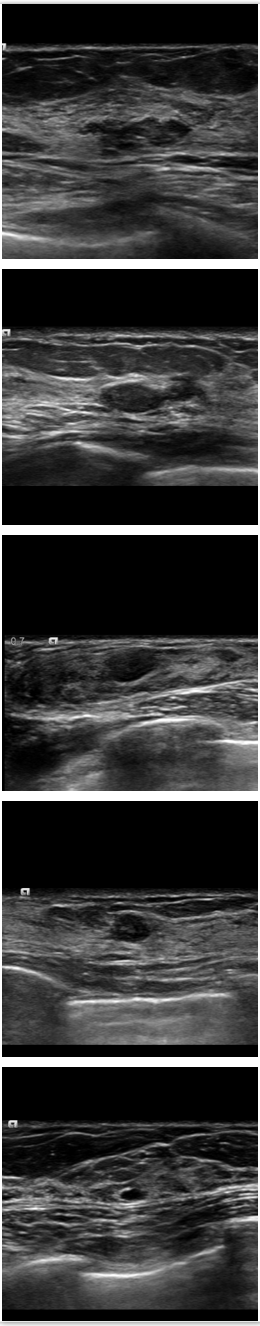}}\hfill
\subfigure[]{\includegraphics[width = .165\linewidth]{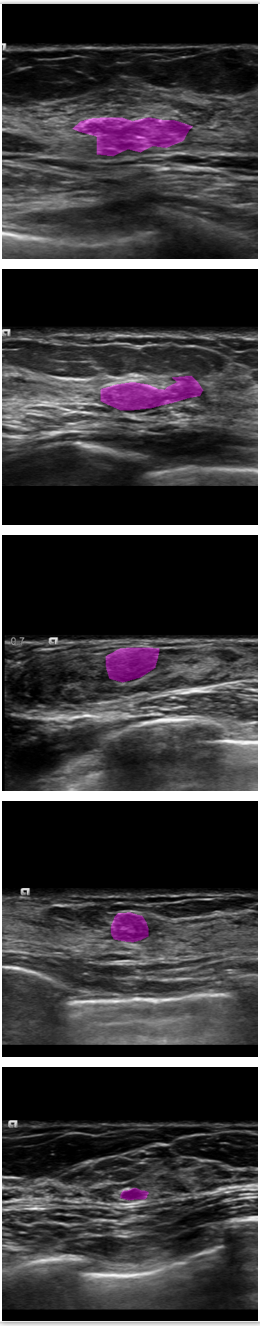}}\hfill
\subfigure[]{\includegraphics[width = .165\linewidth]{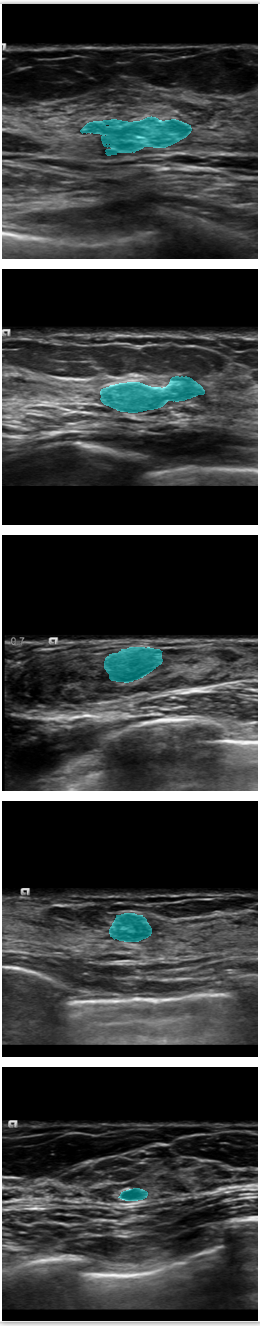}}\hfill
\subfigure[]{\includegraphics[width = .165\linewidth]{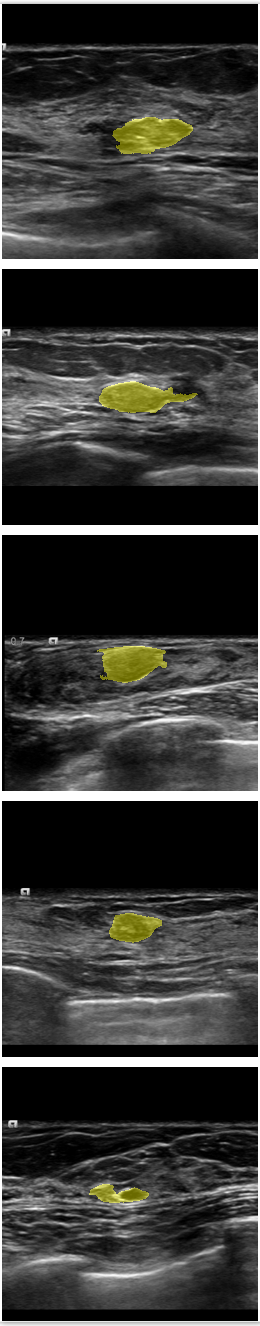}}\hfill
\subfigure[]{\includegraphics[width = .165\linewidth]{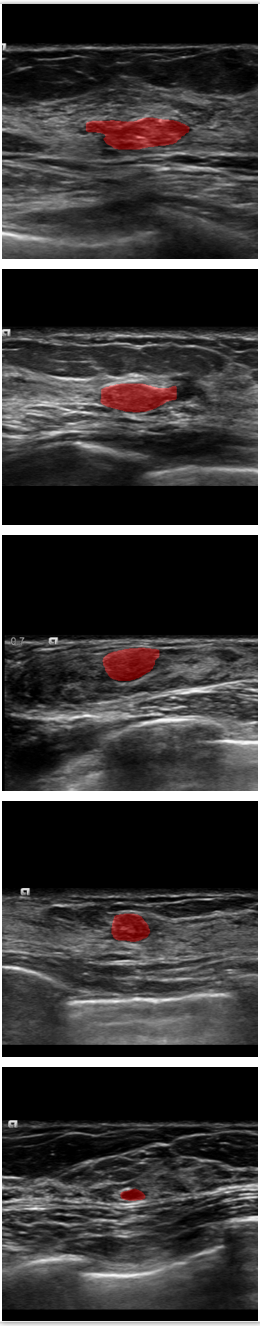}}\hfill
\subfigure[]{\includegraphics[width = .165\linewidth]{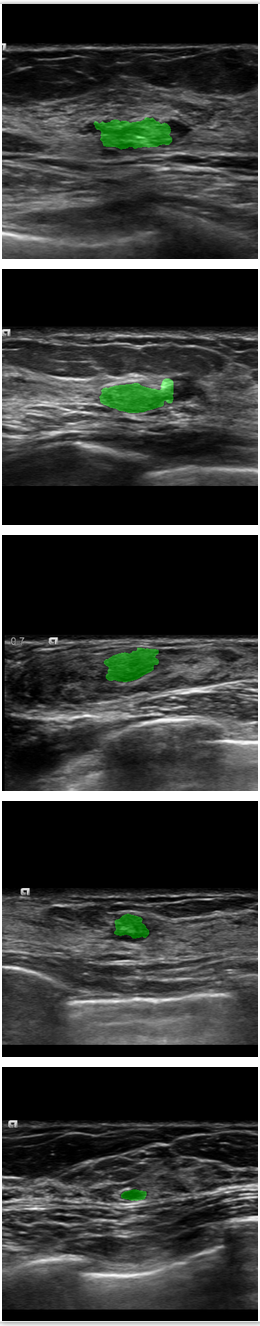}}
\caption{ {Comparison of SPCGAN and other segmentation methods of benign lesions. (a) shows original image of benign lesions, (b) shows the manual annotation, (c) shows the result of SPCGAN ,(d), (e) and (f) show results from ResNet,  Mask R-CNN and level set.}}
\label{fig:benign}
\end{figure}

\subsection{Comparisons among SPCGAN, FCN(ResNet), Mask R-CNN, and Level Set}

\begin{table}[htbp]
{\caption{DSC of different segmentation methods.}}
\centering
\begin{tabular}{ccccc}
\hline
\hline
 & SPCGAN & FCN(ResNet) & Mask R-CNN & level set\\
\hline
All lesions & 0.92$\pm$0.04 & 0.90$\pm$0.07 &  0.89$\pm$0.10 & 0.79$\pm$0.17 \\
benign & 0.92$\pm$0.04 & 0.90$\pm$0.07 &  0.89$\pm$0.10 & 0.83$\pm$0.16 \\
malignant & 0.93$\pm$0.04 & 0.90$\pm$0.07&  0.89$\pm$0.08 & 0.65$\pm$0.17 \\
\hline
\hline
\end{tabular}
\label{tab:alldice}
\end{table}

To evaluate the effect of SPCGAN framework on the breast ultrasound lesion segmentation accuracy, we compared it with FCN(ResNet) framework, Mask R-CNN framework and the traditional segmentation method level set. Table \ref{tab:alldice} summarizes the DSC of different methods on our test database of 141 lesions from the entire dataset and 32 lesions are malignant. 
DSC values were obtained from 109 benign and 32 malignant breast lesions. {
Comparing the overall results of 4 different methods, we see that SPCGAN performed better by 2\% improvement compared to the FCN(ResNet) method (p\textless0.001) and by 3\% improvement compared to the Mask R-CNN method(p\textless0.001). The traditional method level set performed significantly worse compared to SPCGAN method (p\textless0.001). Furthermore, compared to the traditional segmentation method level set, the DSCs obtained from SPCGAN, FCN(ResNet) and Mask R-CNN still remain high no matter a lesion is benign or malignant. The DSCs of malignant lesions from the level set method were much lower than the DSCs of benign lesions.}

{Fig.\ref{fig:benign} displays the segmentation results of our SPCGAN, FCN(ResNet), Mask R-CNN and the level set method from benign lesions. Compared with the FCN(ResNet) (d), Mask R-CNN (e) and the level set (f) method, the results of our SPCGAN (c) show good agreements with the manual contours of the lesions. The segmentations from SPCGAN are very close to manual segmentations.}

\begin{figure}[htbp]
\centering
\subfigure[]{\includegraphics[width = .165\linewidth]{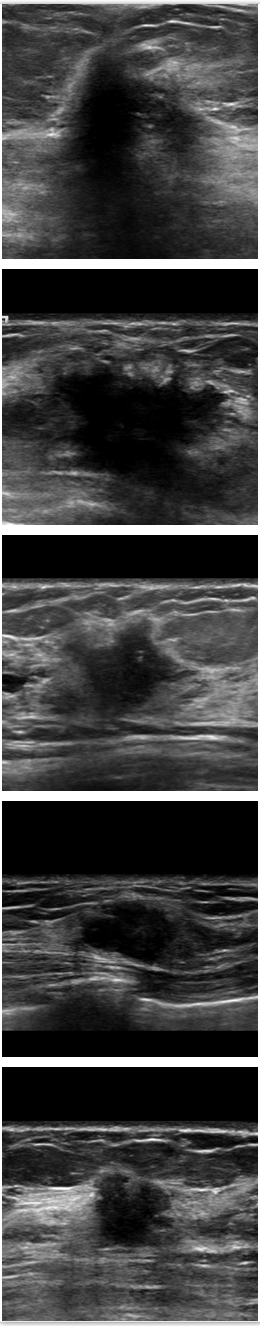}}\hfill
\subfigure[]{\includegraphics[width = .165\linewidth]{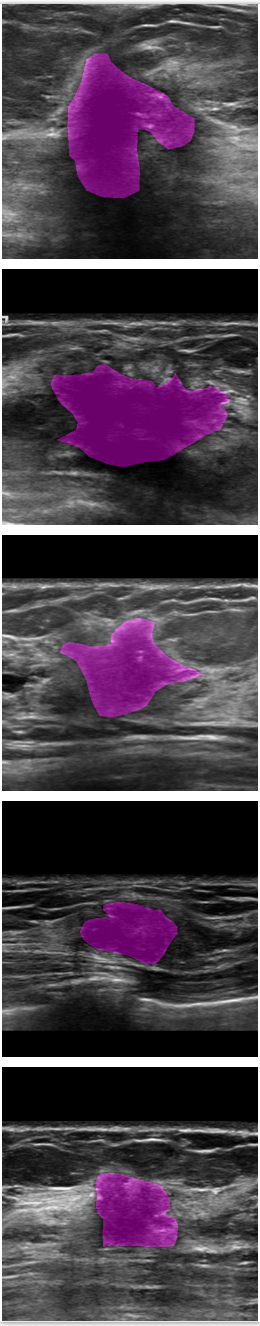}}\hfill
\subfigure[]{\includegraphics[width = .165\linewidth]{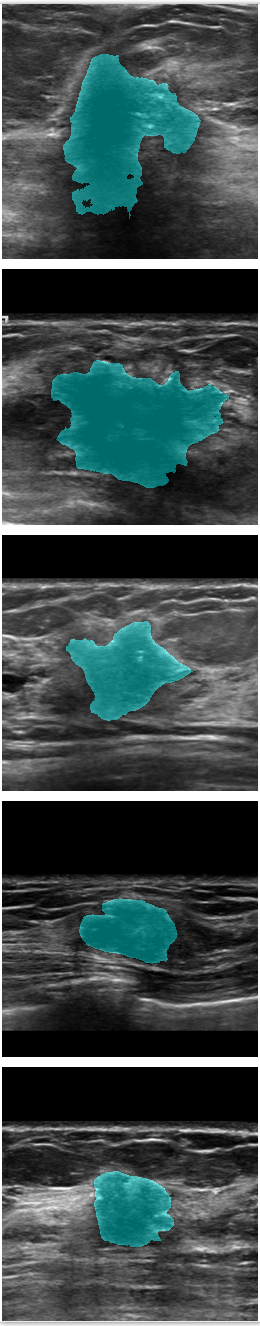}}\hfill
\subfigure[]{\includegraphics[width = .165\linewidth]{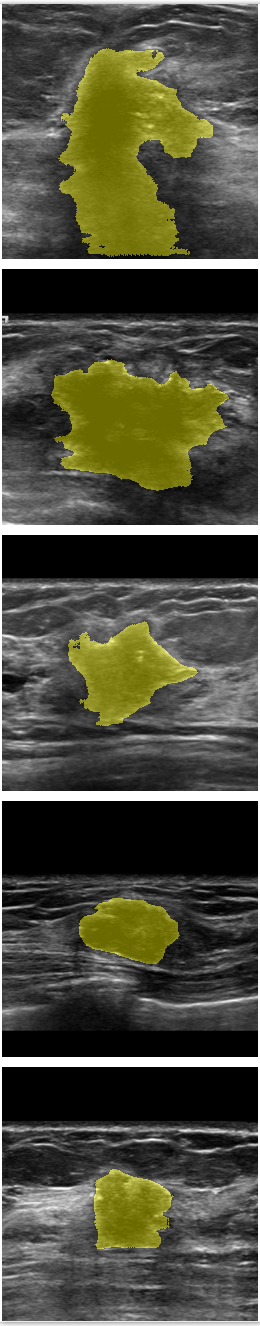}}\hfill
\subfigure[]{\includegraphics[width = .165\linewidth]{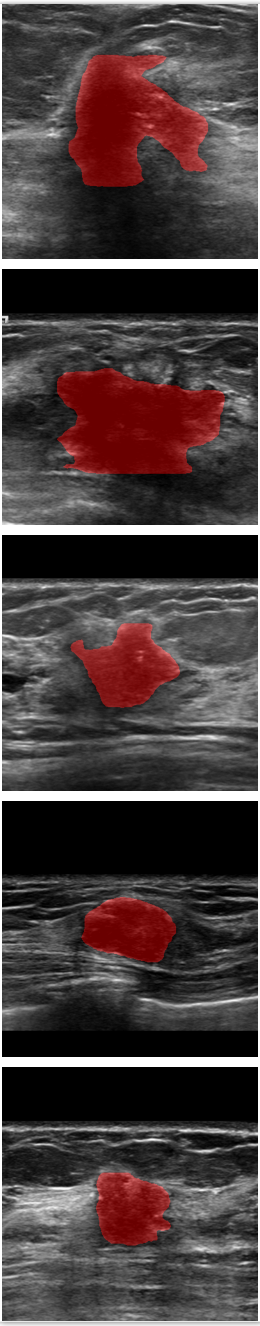}}\hfill
\subfigure[]{\includegraphics[width = .165\linewidth]{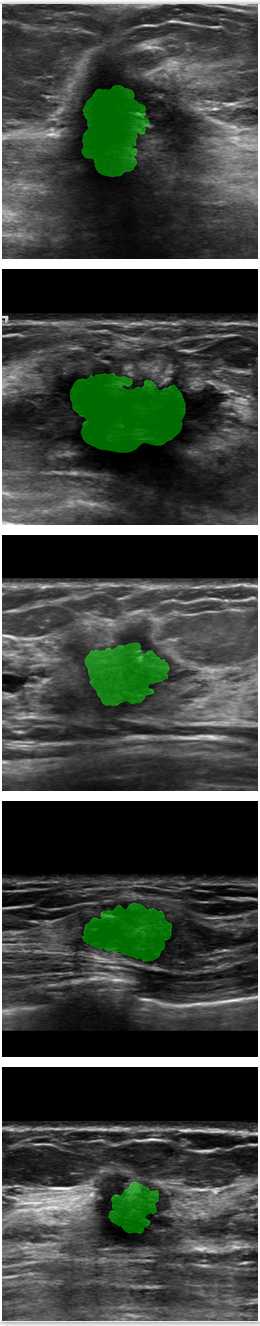}}
\caption{ {Comparison of SPCGAN and other segmentation methods of malignant lesions.(a) shows original image of malignant lesions, (b) shows the manual annotation, (c) shows the result of SPCGAN ,(d), (e) and (f) show results from FCN(ResNet), Mask R-CNN and the level set method.}}
\label{fig:malignant}
\end{figure}

{The examples given in Fig.\ref{fig:malignant} correspond to the segmentation results of our SPCGAN, FCN(ResNet), Mask R-CNN and the level set method from malignant lesions. The FCN(ResNet) tends to oversegment the cancer when there is posterior shadowing, especially for the lesion in the first row. SPCGAN shows relatively more robust performance compared to FCN(ResNet), Mask R-CNN and the level set method.}

\begin{figure}[ht]
\centering
\includegraphics[scale=0.2]{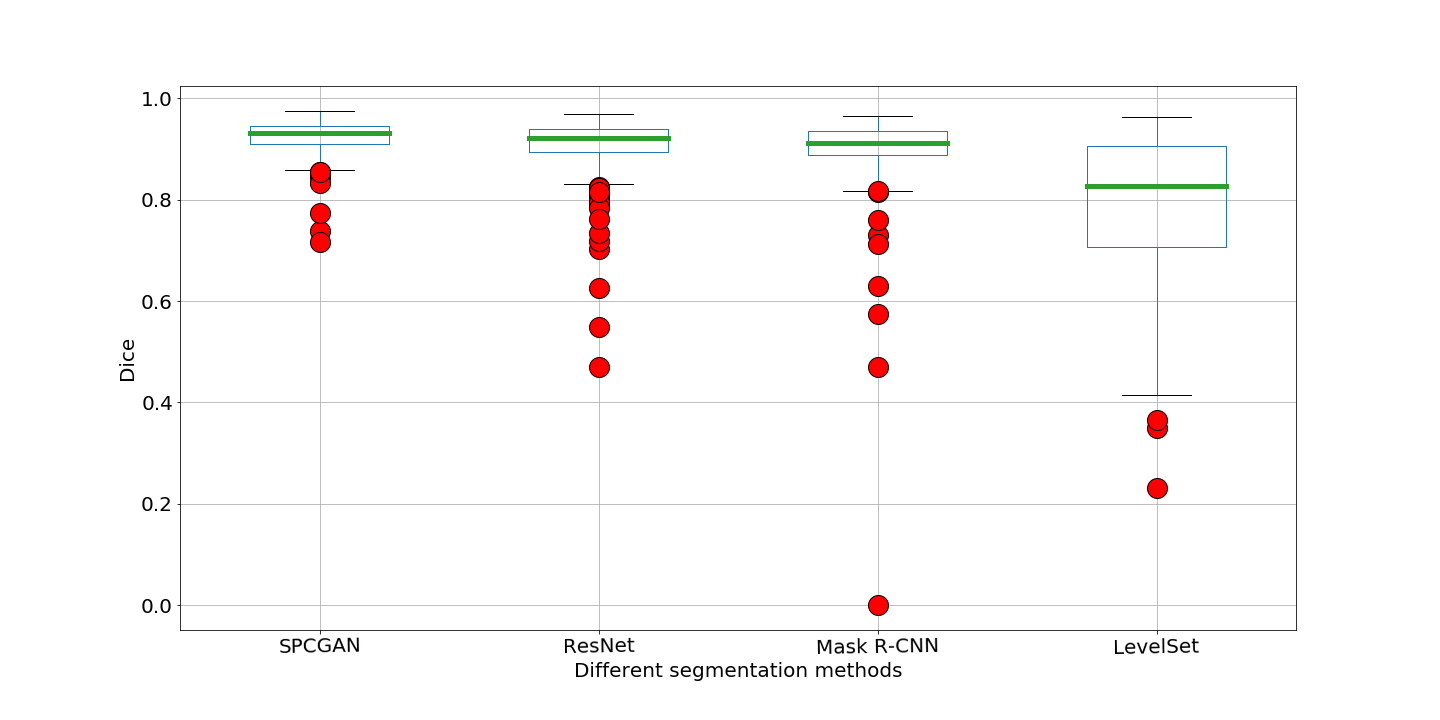}
\caption{{Boxplot of DSC from different segmentation methods with SPCGAN  using backbone of ResNet.}}
\label{fig:boxplot}
\end{figure}

Fig.\ref{fig:boxplot} illustrates boxplots of DSC for different segmentation methods. We can see that results from our SPCGAN have much less variance compared to other methods.

\subsection{Comparisons between Frameworks Trained with Varying Number of Samples}

\begin{table}[htbp]
\caption{DSC of different segmentation frameworks using backbone of ResNet.}
\centering
\begin{tabular}{cccc}
\hline
\hline
number of training samples & SPCGAN & GAN & FCN \\
\hline
20 & 0.79$\pm$0.26 & 0.74$\pm$0.30 & 0.64$\pm$0.37 \\
40 & 0.85$\pm$0.20 & 0.84$\pm$0.22 & 0.83$\pm$0.23 \\
80 & 0.88$\pm$0.10 & 0.86$\pm$0.20 & 0.87$\pm$0.15  \\
200 & 0.91$\pm$0.08 & 0.90$\pm$0.08 & 0.90$\pm$0.09  \\
399 & 0.92$\pm$0.04 & 0.91$\pm$0.07 & 0.90$\pm$0.07 \\
\hline
\hline
\label{tab:differentsize_resnet}
\end{tabular}
\end{table}

\begin{figure}[ht]
\centering
\includegraphics[scale=0.5]{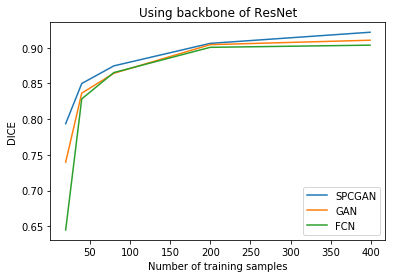}
\caption{DSC values obtained when SPCGAN ,GAN and FCN were trained with different number of training samples based on ResNet.}
\label{fig:plot_resnet}
\end{figure}

In order to compare the performance of SPCGAN, GAN and FCN(ResNet) model, we use varying numbers of samples to train the model and then test with the same testing dataset. The changes in the performance of SPCGAN, GAN and FCN(ResNet) when trained with 20, 40, 80, 200 and 399 samples are displayed in Fig. \ref{fig:plot_resnet}.

From Fig. \ref{fig:plot_resnet} and Table \ref{tab:differentsize_resnet}, we can observe that SPCGAN model obtained best results among all training sample numbers, especially when sample size was small. Although GAN and FCN(ResNet) model trained with 200 samples can achieve DSC of 0.90, it is still 0.01 lower than SPCGAN. Particularly, when training samples increased to 399, the DSC of SPCGAN improved to 92\% while that of FCN(ResNet) remained with 90\%. The performance of GAN is between that of SPCGAN and FCN(ResNet).

\begin{figure}[htbp]
\centering
\subfigure[]{\includegraphics[width = .24\linewidth]{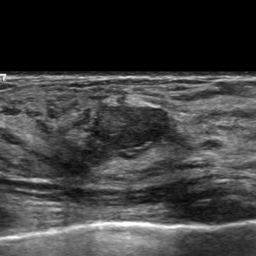}}\hfill
\subfigure[]{\includegraphics[width = .24\linewidth]{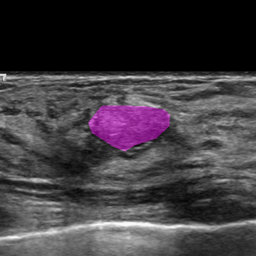}}\hfill
\subfigure[]{\includegraphics[width = .24\linewidth]{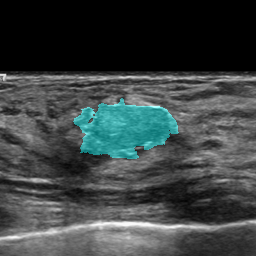}}\hfill
\subfigure[]{\includegraphics[width = .24\linewidth]{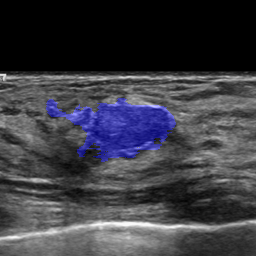}}\hfill
\subfigure[]{\includegraphics[width = .24\linewidth]{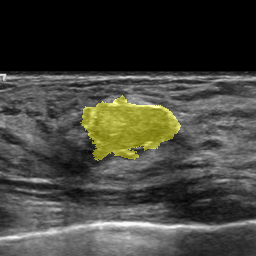}}\hfill
\subfigure[]{\includegraphics[width = .24\linewidth]{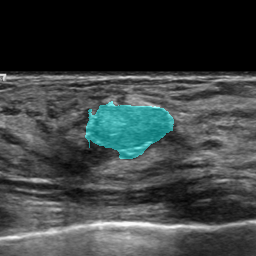}}\hfill
\subfigure[]{\includegraphics[width = .24\linewidth]{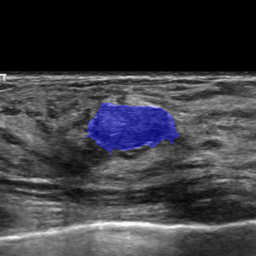}}\hfill
\subfigure[]{\includegraphics[width = .24\linewidth]{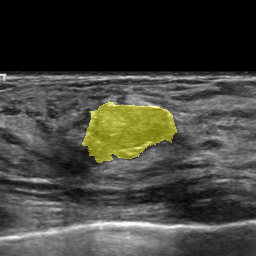}}\hfill
\subfigure[]{\includegraphics[width = .24\linewidth]{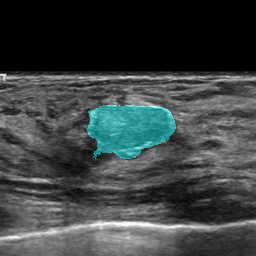}}
\subfigure[]{\includegraphics[width = .24\linewidth]{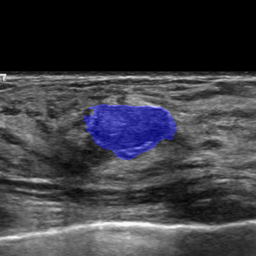}}
\subfigure[]{\includegraphics[width = .24\linewidth]{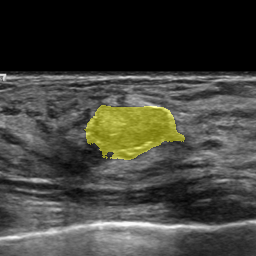}}
\caption{ A breast lesion, which was segmented by SPCGAN, GAN and FCN(ResNet) trained with varying number of samples. (a) shows the original image, (b) shows the manual annotation, (c) shows the result of SPCGAN with 20 training samples, (d) shows the result of GAN with 20 training samples, (e) shows the result of FCN(ResNet) with 20 training samples, (f) shows the result of SPCGAN with 200 training samples, (g) shows the result of GAN with 200 training samples, (h) shows the result of FCN(ResNet) with 200 training samples, (i),(j) and (k) show result of SPCGAN,GAN and FCN(ResNet) with 399 training samples respectively.}
\label{fig:differentsize}
\end{figure}

Fig. \ref{fig:differentsize} displays one case for which segmentation was performed by SPCGAN and FCN(ResNet) trained with varying numbers of samples. This example demonstrates how unclear boundary and shadow in ultrasound images may affect segmentation algorithms.

%dice0.895424	0.814133	0.880784	0.810763  0.853754	0.814938

\subsection{Results from Test Data from Other Manufactures}

\begin{figure}[htbp]
\centering
\subfigure[]{\includegraphics[width = .19\linewidth]{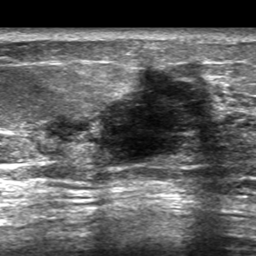}}\hfill
\subfigure[]{\includegraphics[width = .19\linewidth]{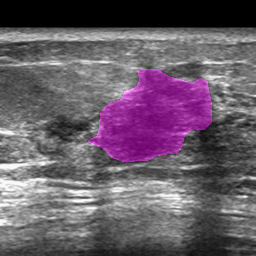}}\hfill
\subfigure[]{\includegraphics[width = .19\linewidth]{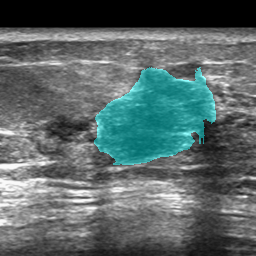}}\hfill
\subfigure[]{\includegraphics[width = .19\linewidth]{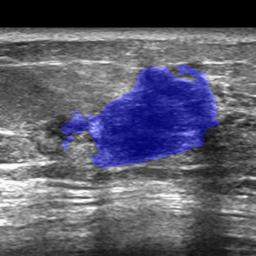}}\hfill
\subfigure[]{\includegraphics[width = .19\linewidth]{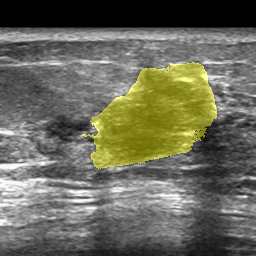}}
\caption{ Comparison among SPCGAN, GAN and FCN(ResNet) in a TOSHIBA image. (a) shows original image of benign lesions, (b) shows the manual annotation, (c) shows the result of SPCGAN ,(d) shows the result of GAN , and (e) shows results from FCN(ResNet).}
\label{fig:toshiba}
\end{figure}

\begin{table}[htbp]
\caption{DSC for TOSHIBA images}
\centering
\begin{tabular}{cccc}
\hline
\hline
 & SPCGAN & GAN & FCN(ResNet)\\% & Mask R-CNN
\hline
30 test images & 0.93$\pm$0.02&  0.92$\pm$0.04   &0.92$\pm$0.04\\% & 0.88$\pm$0.15
\hline
\hline
\end{tabular}
\label{tab:toshiba1}
\end{table}

To explore the performance of SPCGAN on the segmentation quality with test data from different manufactures, we collected 30 BUS images of breast disease scanned from TOSHIBA Ultrasound System and applied our model which was trained on SIEMENS images only.

From Table \ref{tab:toshiba1}, we can observe that the difference between DSC values of SPCGAN and FCN(ResNet) was not statistically significant (p=0.14 with paired t-test). The example given in Fig. \ref{fig:toshiba} corresponds to a breast cancer with the ill-defined boundary. They are both very robust but the mean DSC from SPCGAN is still higher.

\subsection{Comparisons among Models Trained with different backbone networks}
\begin{figure}[ht]
\centering
\includegraphics[scale=0.5]{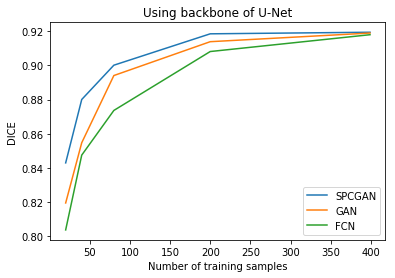}
\caption{DSC values obtained when SPCGAN ,GAN and FCN were trained with different number of training samples based on U-Net.}
\label{fig:plot_unet}
\end{figure}

In this section , we repeated the experiments where instead of using a ResNet method in the FCN part we use a U-Net method in the FCN part. From Fig. \ref{fig:plot_unet} and Table \ref{tab:differentsize_unet}, we can observe that SPCGAN model still obtained best results among all training sample numbers. Table \ref{tab:differentsize2} and Table \ref{tab:toshiba2} show the DSC of three different frameworks (SPCGAN, GAN and backbone structure only) based on two different backbone structures (ResNet and U-Net) on Siemens and Toshiba datasets. SPCGAN frame archives the best performance.

\begin{table}[htbp]
\caption{DSC of different segmentation frameworks using backbone of U-Net.}
\centering
\begin{tabular}{cccc}
\hline
\hline
number of training samples & SPCGAN & GAN & FCN \\
\hline
20 & 0.84$\pm$0.17 & 0.82$\pm$0.18 & 0.80$\pm$0.22 \\
40 & 0.88$\pm$0.10 & 0.85$\pm$0.15 & 0.5$\pm$0.15 \\
80 & 0.90$\pm$0.07 & 0.89$\pm$0.10 & 0.87$\pm$0.12 \\
200 & 0.92$\pm$0.05 & 0.91$\pm$0.06 & 0.91$\pm$0.06 \\
399 & 0.92$\pm$0.05 & 0.92$\pm$0.04 & 0.92$\pm$0.05 \\
\hline
\hline
\label{tab:differentsize_unet}
\end{tabular}
\end{table}

\begin{table}[htbp]
\caption{DSC for SIEMENS images tested by different segmentation Models Trained with different backbone networks.}
\centering
\begin{tabular}{cccc}
\hline
\hline
 & SPCGAN & GAN & backbone only \\
\hline
ResNet & 0.92$\pm$0.04 & 0.91$\pm$0.06 &0.90$\pm$0.07 \\
U-Net & 0.92$\pm$0.04 & 0.92$\pm$0.04 & 0.92$\pm$0.05 \\
\hline
\hline
\label{tab:differentsize2}
\end{tabular}
\end{table}

\begin{table}[htbp]
\caption{DSC for TOSHIBA images tested by different segmentation Models Trained with different backbone networks.}
\centering
\begin{tabular}{cccc}
\hline
\hline
 & SPCGAN & GAN & backbone only \\ 
\hline
ResNet & 0.93$\pm$0.02&  0.92$\pm$0.04   &0.92$\pm$0.04  \\
U-Net & 0.94$\pm$0.02 & 0.94$\pm$0.02 & 0.93$\pm$0.02 \\
\hline
\hline
\end{tabular}
\label{tab:toshiba2}
\end{table}

\section{Conclusion and Discussion}
In this study, we proposed a CycleGAN based model for segmenting breast lesions in 2D breast ultrasound. We compared our model with FCN and the level set based approach. The results show that our model is the most robust and accurate with a DSC of 0.92$\pm$ 0.04. Without retraining, the same model is applied on ultrasound images from a different manufacture, resulting a DSC of 0.93$\pm$ 0.02.

The novelty of our work is the combination of the CycleGAN and the pixel-wise cost which makes the model has the advantage of both GAN and FCN. The similar trends are observed for different frameworks with two different backbone structures (ResNet and U-Net). {However, for some challenging images, especially when calcification is present, the segmentation is still not very smooth. We believe more calcification cases in the training samples will better improve the performance.} From Fig. \ref{fig:plot_resnet}, we can observe an improvement on DSC from 0.79 to 0.92 when the number of training images is increased. 
% * <byw94@hotmail.com> 2018-11-24T05:02:26.101Z:
% 
% we will increase the size of our dataset in order to obtain a more robust model.
% 
% ^.
Another approach is to apply post processing, for example, Markov random filtering, to make segmentation smooth and complete. Both deep learning based methods are significantly better than the traditional level set approach in both malignant and benign cases, even when the deep learning model is only trained with 20\% percent of the dataset. 
With more annotated data, the performance of supervised learning can be improved significantly.  
% * <byw94@hotmail.com> 2018-11-24T05:04:39.829Z:
% 
% t 
% 
% ^.
It is still possible to enhance the segmentation performance by combining deep learning methods  with geometric methods that take into account the context of local orientations (in the images and/or segmentation boundaries) via group-CNNs \cite{bekkers2018roto-translation} instead of the normal CNNs (FCNs) used in this paper.  

%%%Questions:
%1.cycle backward 的generator output是什么-> 从Segmentation生成原始图像
%2.backward generator的input是forward生成的Segmentation还是ground truth的Segmentation？
%3.为什么不用还原回去的原始图像也做一次pixel-wise的loss calculation然后和真正的原始图像比较？
%4.为什么要用cycleGAN？不能单独用一个GAN,然后直接在discriminator那里做像素级的判别吗？
In the cyclic training process of our model, the forward generator firstly produces automatic segmentation of the breast lesions indistinguishable to the forward discriminator by minimizing the adversarial loss, cycle consistency loss and pixel-wise cost. This generated segmentation is then fed into the backward generator which tries to get it recover to the original image. During this stage, the pixel-wise cost is no longer applied as it is hard to recover the original image from the segmentation at a pixel level. The implementation of CycleGAN algorithm effectively utilizes the prior knowledge of lesion images to provide the prior distribution in the latent space for the input. By imposing this prior probability distribution, the mapping between input data sample and real data is under a more sensible constraint. In ultrasound annotation tasks, the segmentation is challenging because of poor quality, posterior shadowing and weak boundaries. In this case, the use of prior knowledge could further improve the robustness of the segmentation.  

Accurate segmentation would help describe shape, orientation, margins, echo pattern, posterior acoustic features, and surrounding tissue alterations of a lesion in BI-RADS US lexicon. The description would also aid radiologists or computer algorithms\cite{Tan2012,Tan2013} to diagnose a lesion. Given accurate segmentations, it would be logical to design further deep learning networks to differentiate malignant lesions from benign lesions or generate BI-RADS scores in ultrasound. 

One limitation of our study is that we compared the segmentation results to annotations from only one medical expert. Although beyond the scope of the present work, it is of interest to include multiple annotations by several medical experts. Moreover, as the CycleGAN model has the ability of learning prior knowledge, for example the shape, it is possible that it only learns the style of one reader. Whether the prior knowledge from one reader is sufficient to obtain good inter-reader variability shall be investigated in the feature. 

In the real clinical practice, there are ultrasound devices from different manufacturers deployed. These images varies in resolution, contrast, and the presence of noise. In this study, we evaluated the possibility of applying our deep learning model trained on SIEMENS ultrasound images only to TOSHIBA ultrasound images. The segmentation accuracy still remains high. Researchers can focus on developing robust algorithms on different types of ultrasound images, which is important to make computer techniques available to real world practice.

In this work we focused on segmenting a specific type of lesions in ultrasound images of breasts. As a result we take advantage both of the GAN (acting globally on full images) and of the FCN (with a pixel-wise loss to account for local optimization) in our unifying machine learning approach for lesion segmentation. Our method has the potential to help radiologists delineate breast lesion and improve the efficiency of workflow for reporting and inter-/intra- reader variability. As we achieved promising results on our two datasets, it is also interesting to apply the technique to other segmentation problems in medical imaging. 
% * <byw94@hotmail.com> 2018-11-24T05:17:09.992Z:
% 
% the workflow of?? inter-/intra- reader variability.
% 
% ^.
While focused on one type of lesions in ultrasound, future work can address a wider range of objects in different medical images.  

\bibliographystyle{IEEEtran}
\bibliography{IEEEabrv,fea_cfshan.bib}
\vspace{-10 mm} 
\begin{IEEEbiography}
[{\includegraphics[width=1in,height=1.25in,clip,keepaspectratio]{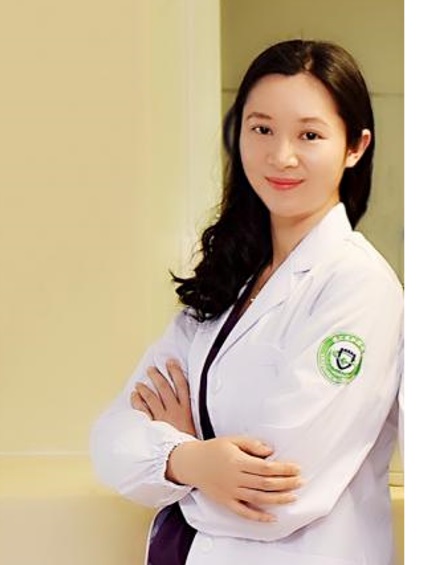}
}]{Jie Xing}
is an Attending doctor with Department of Gynecological oncology surgery, Institute of Cancer Research and Basic Medical Sciences of Chinese Academy of Sciences, Cancer hospital of Chinese Academy of Sciences, Zhejiang Cancer Hospital.Her main expertise are surgical treatment of benign and malignant tumors, chemotherapy immunotherapy and targeted therapy of gynecological tumors. Her research interests are medical artificial intelligence and medical big data.
\vspace{-10 mm}
\end{IEEEbiography}

\begin{IEEEbiography}
 [{\includegraphics[width=1in,height=1.25in,clip,keepaspectratio]{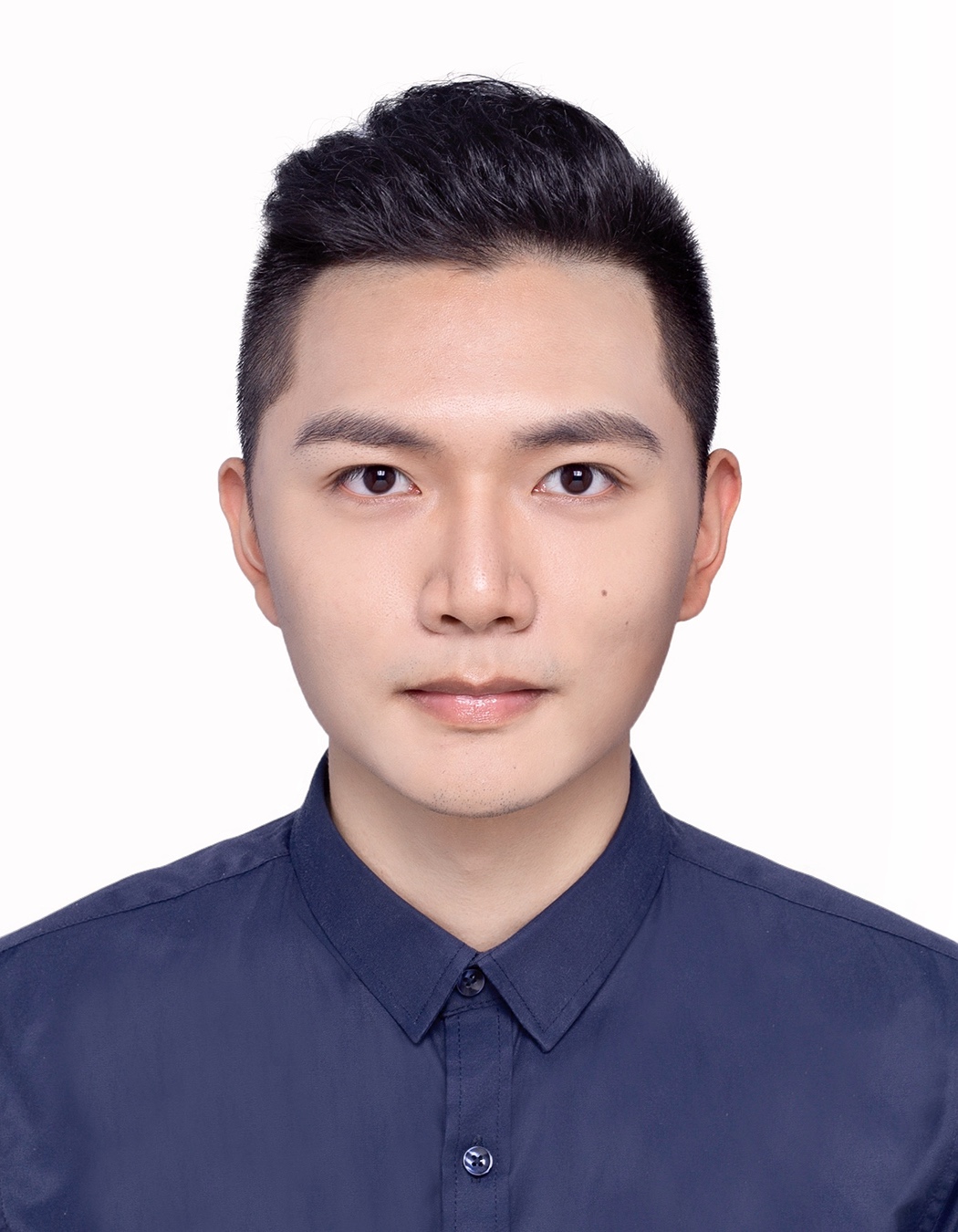}}]{Zheren Li}
received his master degree from the Department of Bioengineering at Imperial College London. He is now a Eng.D student of the School of Biomedical Engineering of Shanghai Jiao Tong University. His research interests are medical image processing and artificial intelligence in medical imaging.
\vspace{-10 mm} 
\end{IEEEbiography}

\begin{IEEEbiography}
 [{\includegraphics[width=1in,height=1.25in,clip,keepaspectratio]{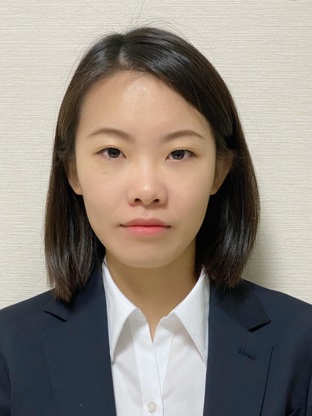}}]{Biyuan Wang}
received her undergraduate degree in the Department of Bioengineering at Imperial College London. She is now a master student of the Department of Computer Science at Tokyo Institute of Technology.
\vspace{-10 mm} 
\end{IEEEbiography}

\begin{IEEEbiography}
 [{\includegraphics[width=1in,height=1.25in,clip,keepaspectratio]{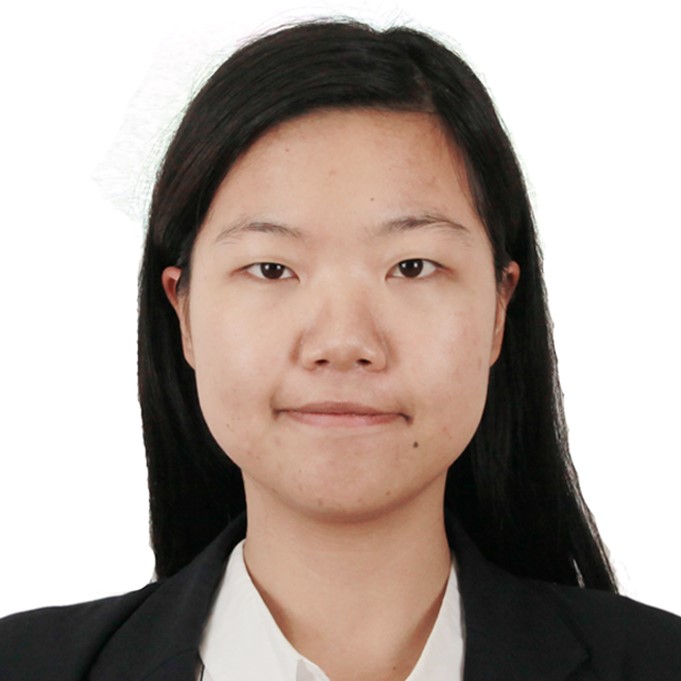}}]{ Yuji Qi}
is a M.S. student from department of Biomedical Engineering, Yale University. Her research interest is artificial intelligence in medical imaging.
\vspace{-140 mm} 
\end{IEEEbiography}

\begin{IEEEbiography}
 [{\includegraphics[width=1in,height=1.25in,clip,keepaspectratio]{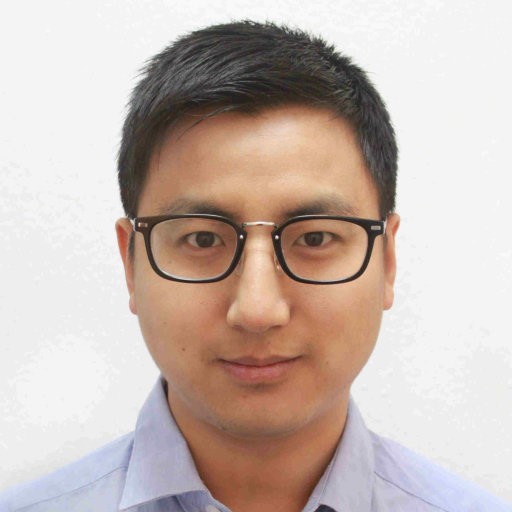}}]{Bingbin Yu}
is a researcher from German Research Center for Artificial Intelligence. His interest is in soft robotics and related machine learning techniques.
\vspace{-140 mm} 
\end{IEEEbiography}

\begin{IEEEbiography}
 [{\includegraphics[width=1in,height=1.25in,clip,keepaspectratio]{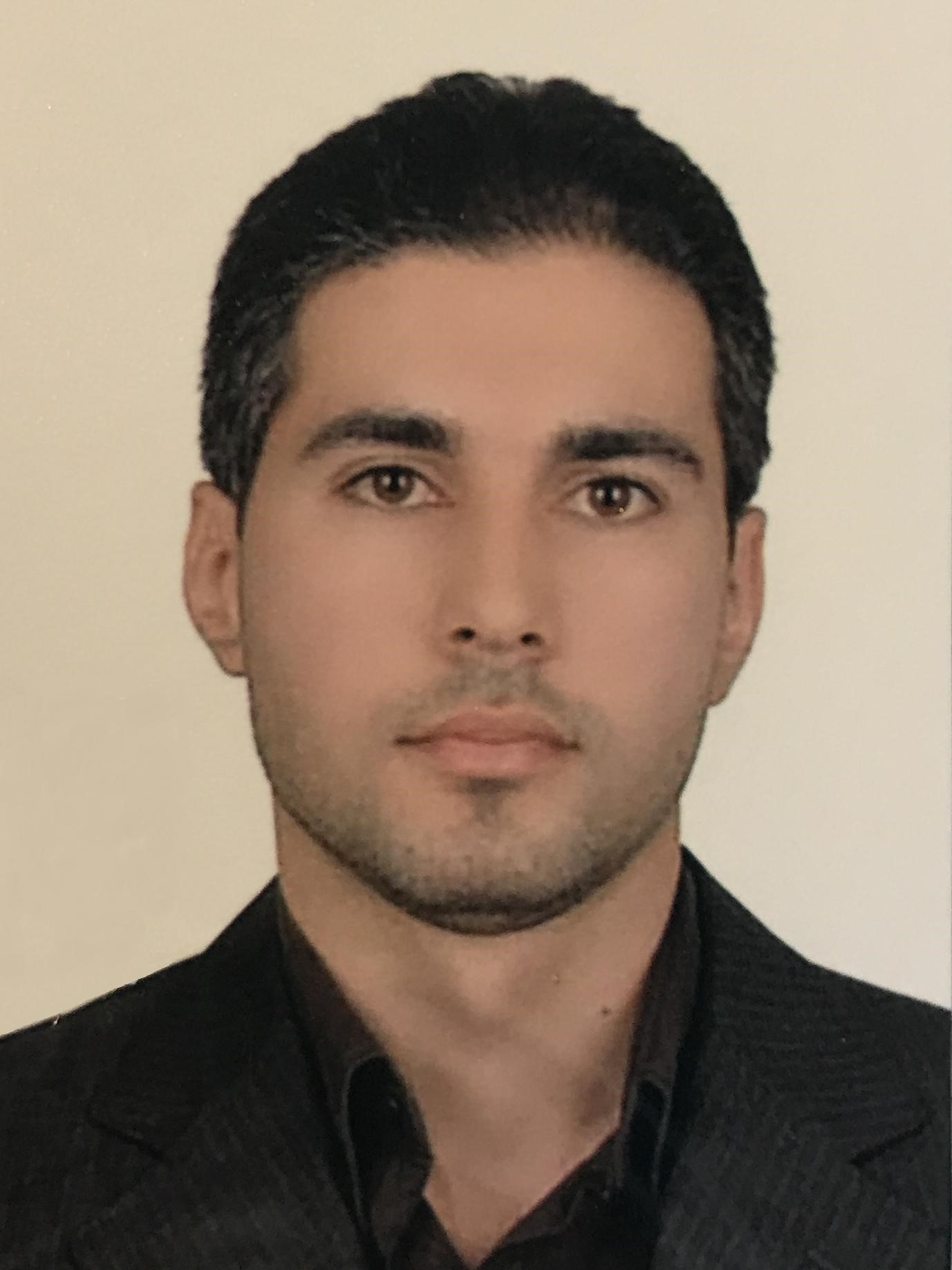}}]{Farhad Ghazvinian Zanjani}
is a PhD student from Department of Electrical Engineering, Center for Care \& Cure Technology Eindhoven, Department of Electrical Engineering, Video Coding \& Architectures.
 
\end{IEEEbiography}

\newpage

\begin{IEEEbiography}
 [{\includegraphics[width=1in,height=1.25in,clip,keepaspectratio]{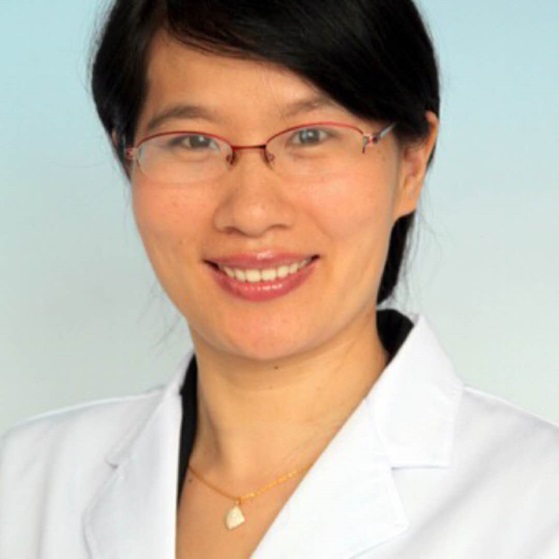}}]{Ai-Wen Zheng}
is a Chief physician with Department of Gynecological oncology surgery, Institute of Cancer Research and Basic Medical Sciences of Chinese Academy of Sciences, Cancer hospital of Chinese Academy of Sciences, Zhejiang Cancer Hospital.She is expert in surgical treatment of gynecological tumors. Her research interests are tumor diagnosis, molecular biology of tumor and medical artificial intelligence.
\vspace{-140 mm} 
\end{IEEEbiography}

\begin{IEEEbiography}
 [{\includegraphics[width=1in,height=1.25in,clip,keepaspectratio]{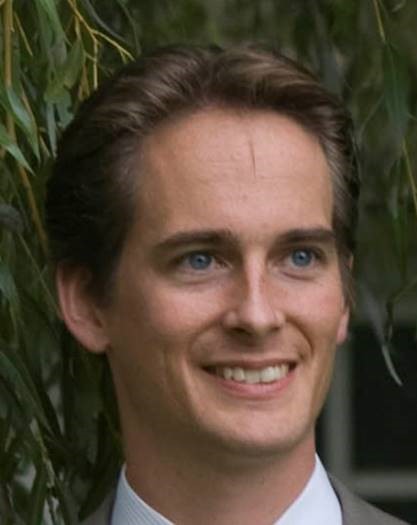}}]{Remco Duits}
is a Mathematician from Eindhoven University of Technology Department of Mathematics and Computer Science. He develops new mathematical theory and algorithms to tackle image analysis applications.His main expertise are differential geometry, harmonic analysis, probability theory, mathematical image analysis and medical Image Analysis. 
\vspace{-140 mm} 
\end{IEEEbiography}

\begin{IEEEbiography}
 [{\includegraphics[width=1in,height=1.25in,clip,keepaspectratio]{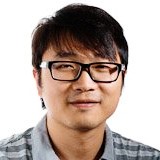}}]{Tao Tan}
is an assistant professor with Department of Mathematics and Computer Science, Centre for Analysis, Scientific Computing, and Applications W\&I,Eindhoven University of Technology and an associated scientific staff with Radiology, Netherlands Cancer Institute. His research interest is computer-aided diagnosis in breast imaging.
\end{IEEEbiography}

\end{document}